\documentclass[11pt,a4paper]{article}

\usepackage[T1]{fontenc}
\usepackage{amsmath,amssymb,amsthm,mathtools}
\usepackage{newtxtext,newtxmath} 

\usepackage{geometry}
\usepackage{booktabs}
\usepackage{longtable}
\usepackage{array}
\usepackage{graphicx}
\usepackage{float} 
\usepackage{hyperref}
\hypersetup{colorlinks=true,linkcolor=blue,citecolor=blue,urlcolor=blue}
\usepackage{enumitem}
\usepackage{caption}

\newcommand{\includeimage}[2][]{%
  \IfFileExists{#2}{%
    \includegraphics[#1]{#2}%
  }{%
    \fbox{\parbox[c][5cm][c]{0.85\textwidth}{\centering \textbf{[Missing Image]} \\ #2}}%
  }%
}

\title{SKG in Practice: OOD Generalization Cases Across Four Domains}
\author{Jian An       \and Chaoshi Wu      \and Xinpan Luo}
\date{October 22, 2026}

\begin{document}
\maketitle

\begin{abstract}
This paper presents the engineering implementation methods of the SKG inference framework and provides complete case studies across four domains.

\begin{table}[H]
\centering
\small
\renewcommand{\arraystretch}{1.3}
\begin{tabular}{p{1.2cm} p{3.5cm} p{9cm}}
\toprule
Case & Domain & Key Results \\
\midrule
1 & 2D Dissipative Mechanical System & OOD MSE is 1.69--4.12 times lower than the best neural baseline; state-dependent metric yields a \(10^8\)-order improvement under controlled conditions \\
2 & Target Detection Sim-to-Real & Real-machine mAP 0.2564 \(\to\) 0.4964 (+24.00 points, single-run experiment, 32 images, point estimate), AP75 0.1736 \(\to\) 0.5765 (+40.29 points), no real-machine retraining required \\
3 & Sequence Modeling sine\(\to\)square & OOD MSE 0.0296 vs Transformer 0.0641 (mean of 5 seeds, p=0.024) \\
4 & Molecular Dynamics MD17 & ID@100 ADE 0.0285 vs ESTAG 0.1467; OOD@100 0.0316 vs 0.1411 (single seed) \\
\bottomrule
\end{tabular}
\end{table}

For each case, we present the objective, applied SKG theory, parameter settings, baseline comparison protocol, data results, and sources of advantage. Following the cases, we provide the complete methodology: pre-validation, task invariant construction, three-stage training, residual training, and the four theorems and one proposition supporting these methods. \textbf{This paper not only demonstrates effective configurations of SKG across four domains but also honestly reports the variance of single-run experiments, limitations of single-seed results, negative updates caused by parameterization mismatch, and practical experience of using engineering thresholds as diagnostic signals rather than hard constraints.}

\end{abstract}

\tableofcontents
\newpage

\section{Introduction}

\subsection{Positioning of This Paper}
This paper is an engineering practice guide for the SKG framework. It targets engineers and researchers who wish to use SKG to solve OOD problems. It answers: what steps to follow to achieve better OOD results than strong existing baselines.

\subsection{Relationship with Paper A \cite{EA2}}
This paper (Paper B) is the engineering implementation companion document to Paper A \cite{EA2}. Paper A establishes the mathematical foundations of SKG (octuple, invariant-induced metric, Fisher objectivity) and rigorous mathematical proofs of the complete learning theory; this paper does not repeat the derivations but focuses on how to transform these theories into operational engineering specifications (Sections 7--11) and verifies their OOD generalization capabilities across four domains (Sections 2--5). Readers seeking theoretical details are referred to Paper A \cite{EA2}. It should be particularly noted that the four theorems and one proposition in Section 10 are independent theoretical contributions of this paper, not repetitions of Paper A.

\subsection{Structure of This Paper}
Sections 2--5 present the four cases, each organized as "Objective \(\to\) Applied SKG Theory \(\to\) Parameter Settings \(\to\) Baseline Comparison Protocol \(\to\) Data Results \(\to\) Sources of Advantage." Section 6 provides a cross-case summary. Sections 7--11 present the complete methodology and supporting theorems. Section 12 concludes.

\section{Case 1: 2D Dissipative Mechanical System}

\subsection{Objective}
To validate the OOD generalization capability of SKG on second-order dissipative systems, compared against neural baselines HNN, NODE, and MLP.

\subsection{Applied SKG Theory}
\begin{itemize}
\item \textbf{Symplectic structure (\(\omega\))}: Constant matrix \(J_c=\begin{pmatrix}0&I\\-I&0\end{pmatrix}\).
\item \textbf{Energy decomposition (\(H, S\))}: \(H=T(p)+V(q)\), dissipation described by \(S\) and \(g\), \(V\) parameterized by Gaussian kernel Nyström.
\item \textbf{State-dependent dissipation metric (\(g\))}: \(G_{pp}(q)\) parameterized by a full symmetric matrix field, softplus ensures positive definiteness.
\item \textbf{Three-stage training}: Trajectory cold start \(\to\) weak-form residual fine-tuning of metric+energy \(\to\) trajectory alignment of decoder.
\item \textbf{Weak-form residual training}: Simpson local integration, avoiding numerical differentiation errors.
\end{itemize}

\subsection{Parameter Settings}
\begin{table}[H]
\centering
\begin{tabular}{ll}
\toprule
Component & Configuration \\
\midrule
Symplectic structure (\(\omega\)) & Constant matrix \\
Mass (\(M_m\)) & Unknown constant positive diagonal, calibrated from training data \\
Potential energy (\(V \subset H\)) & Gaussian kernel Nyström, 96 centers \\
Dissipation (\(g\)) & Full symmetric matrix field, Gaussian kernel Nyström, softplus \\
Training & Weak-form residual, LBFGS, 500 iterations \\
Regularization & Ridge \(10^{-6}\) + Nyström spectral truncation \\
Integration & Strang splitting \\
\bottomrule
\end{tabular}
\end{table}

\textbf{System}:
\begin{equation}
\dot q=M_m^{-1}p,\qquad \dot p=-\nabla_qV(q)-C(q)M_m^{-1}p,
\end{equation}
\begin{equation}
V(q)=\tfrac12q^\top Kq+\tfrac14\|q\|^4,\qquad K=\begin{pmatrix}4&-1\\-1&9\end{pmatrix},
\end{equation}
damping is diagonal in the radial/tangential basis, with both linear and nonlinear function families.

\textbf{Data}: RK4 numerical integration, 20 training trajectories, 12 ID, 12 OOD (initial position and momentum amplitude \(\times 2\)). Time domain 200 steps, \(T=0.2\).

\subsection{Baseline Comparison Protocol}
\begin{itemize}
\item Same training data, same test set, same evaluation metric (200-step MSE).
\item Baselines: HNN, NODE, MLP, each trained under its own paradigm.
\item Ground truth is used only for data generation and evaluation, not as learning input.
\end{itemize}

\subsection{Results}
\begin{figure}[H]
\centering
\includeimage[width=0.9\textwidth]{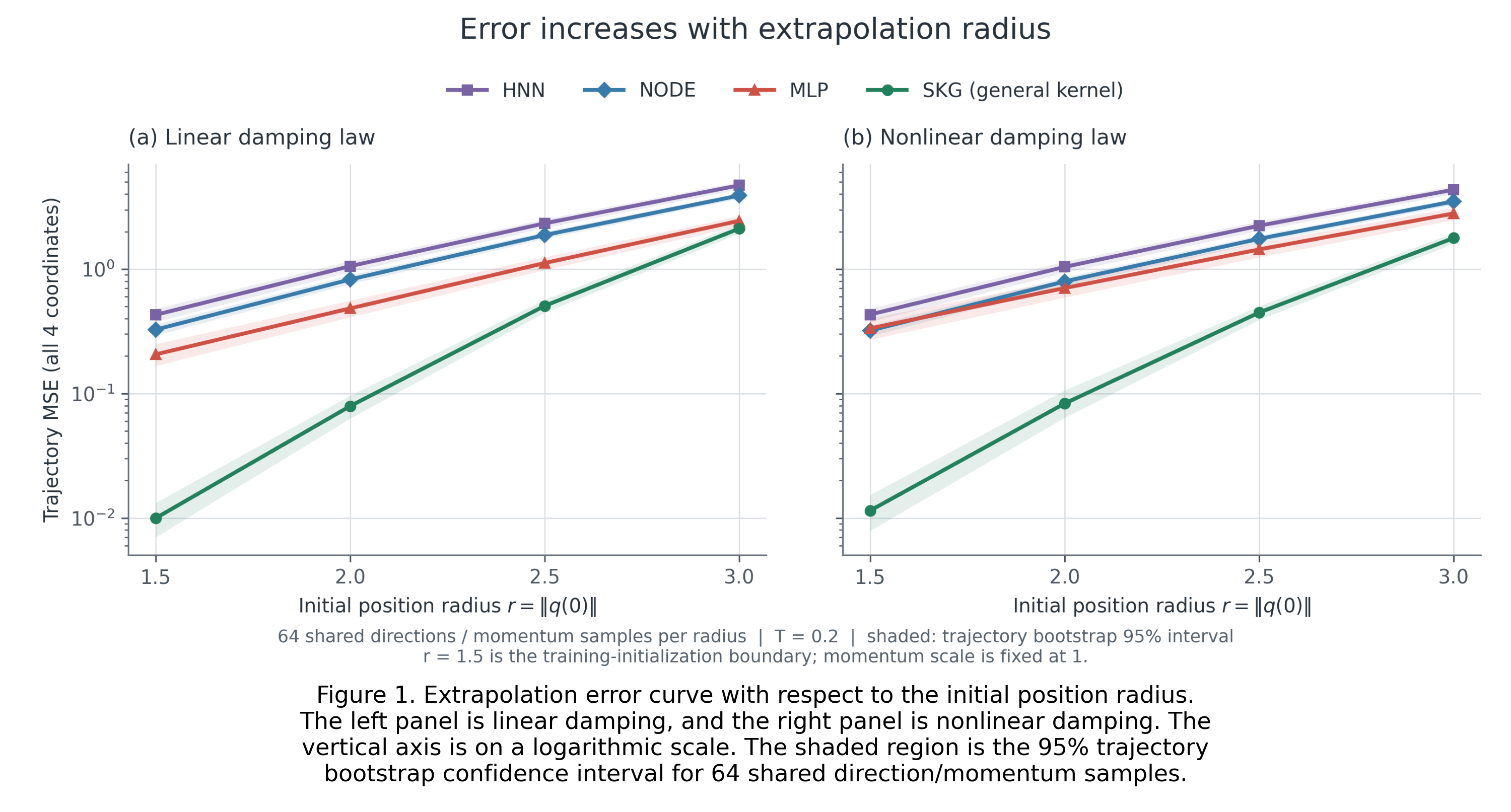}
\refstepcounter{figure}
\label{fig:case1_extrap}
\end{figure}

\textbf{Extrapolation error trend}:
Figure \ref{fig:case1_extrap} shows the trajectory MSE of various methods as the initial position radius increases (extrapolating from the training boundary \(1.5\) to \(3.0\)). Neural network baselines (MLP, NODE, HNN) exhibit exponentially increasing errors with extrapolation distance, while the SKG (General Kernel) error curve grows extremely gently. At the maximum extrapolation radius, SKG's error is 1.69--4.12 times lower than the best baseline.

\begin{figure}[H]
\centering
\includeimage[width=0.9\textwidth]{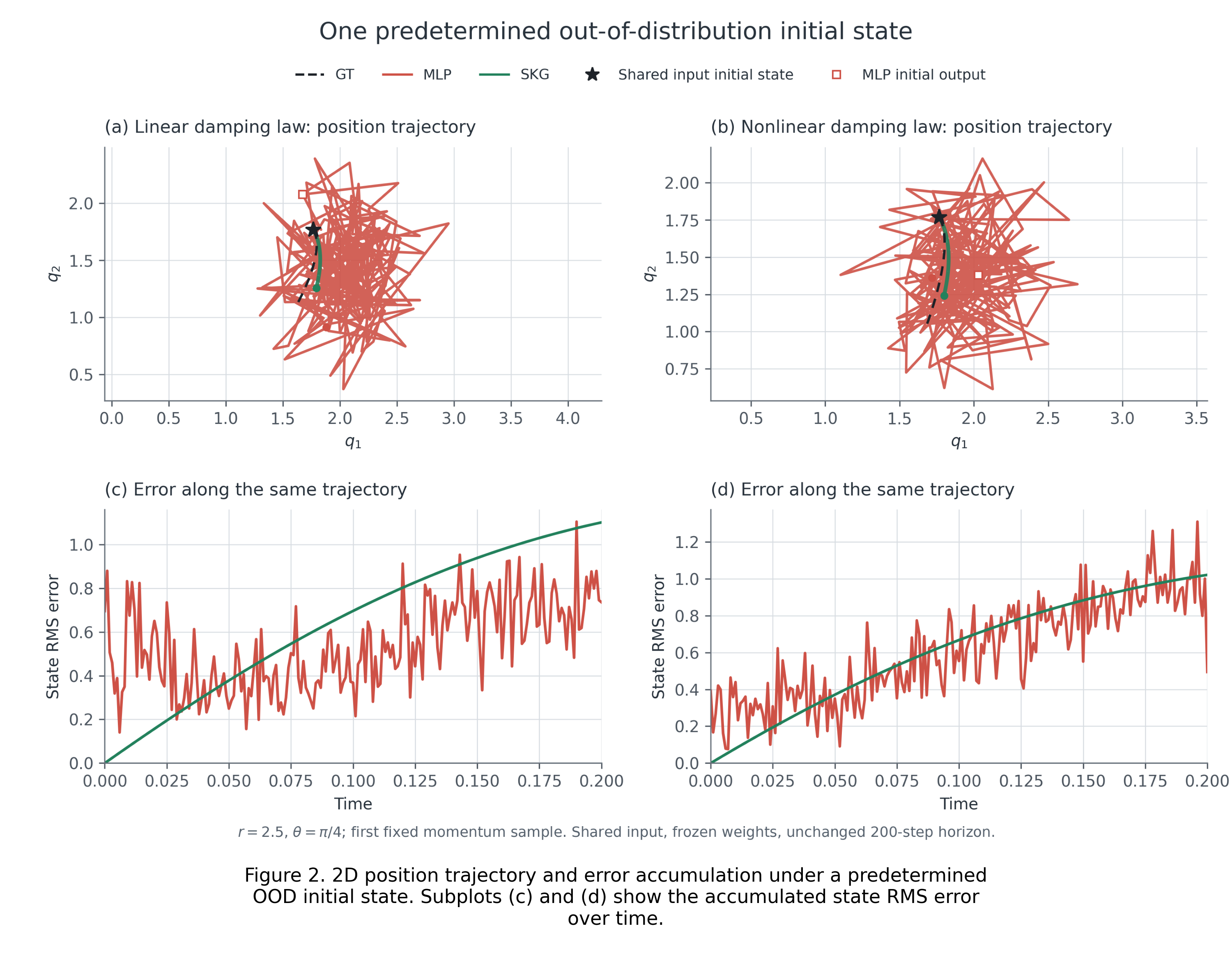}
\refstepcounter{figure}
\label{fig:case1_traj}
\end{figure}

\textbf{2D phase trajectory comparison}:
Figure \ref{fig:case1_traj} shows the position trajectory evolution over 200 steps for a predetermined OOD initial state (\(r=2.5\), outside the training distribution). Lacking physical law constraints, the MLP trajectory (red line) oscillates violently in phase space and completely diverges; while SKG (green line) tightly follows the true trajectory (black dashed line), demonstrating extremely strong dynamical stability.

\begin{figure}[H]
\centering
\includeimage[width=0.8\textwidth]{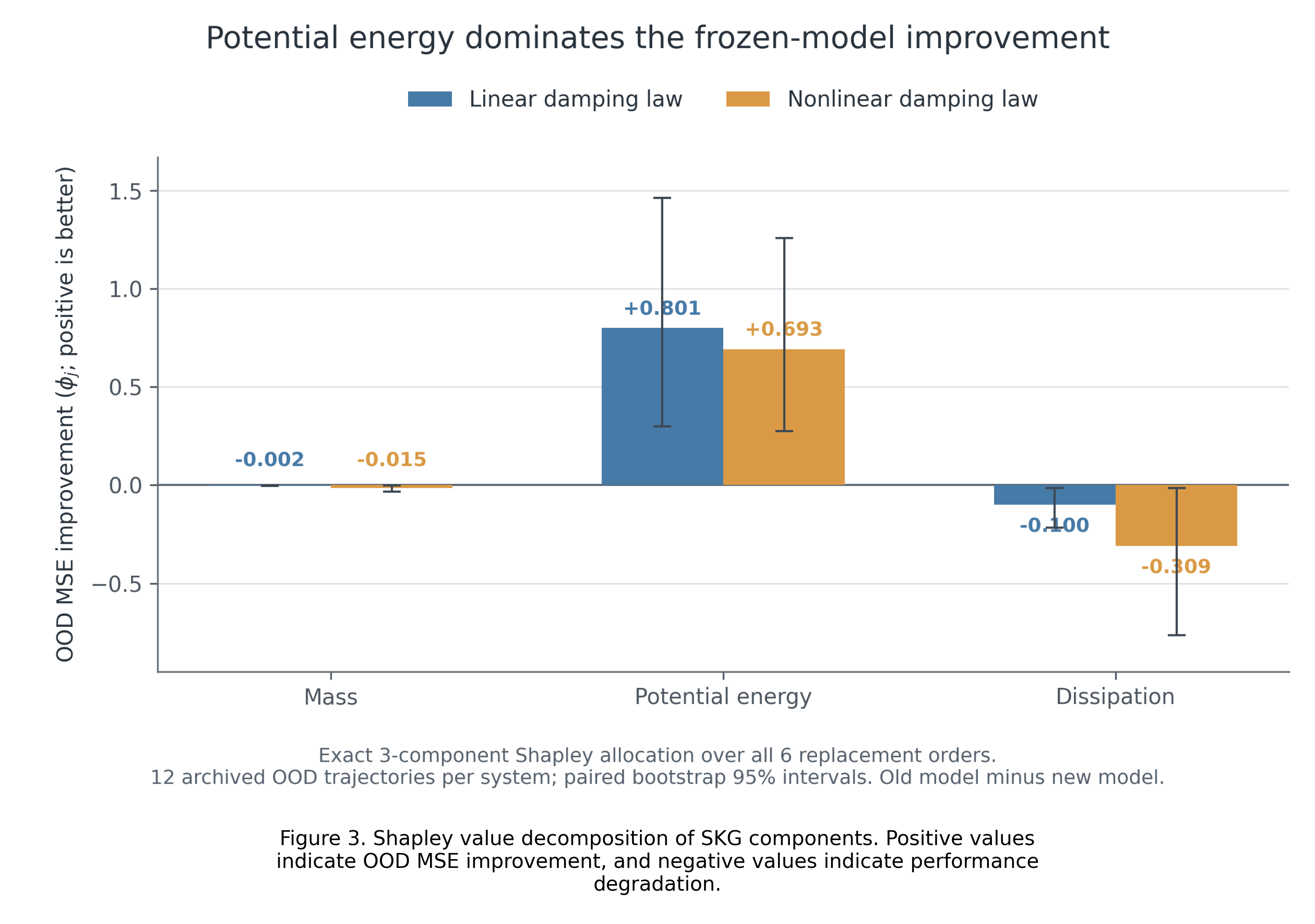}
\refstepcounter{figure}
\label{fig:case1_shapley}
\end{figure}

\textbf{Component ablation (Shapley decomposition)}:
To quantify the contribution of each module within SKG, we performed an exact Shapley decomposition over the mass, potential energy, and dissipation components (Figure \ref{fig:case1_shapley}). The results show that \textbf{the new potential energy is the absolute source of positive gain} (linear +0.801, nonlinear +0.693); the dissipation module instead introduces negative cancellation (linear -0.100, nonlinear -0.309); the contribution of the mass correction is close to zero.

\subsection{Controlled Ablation of the Metric Component}
\begin{figure}[H]
\centering
\includeimage[width=0.8\textwidth]{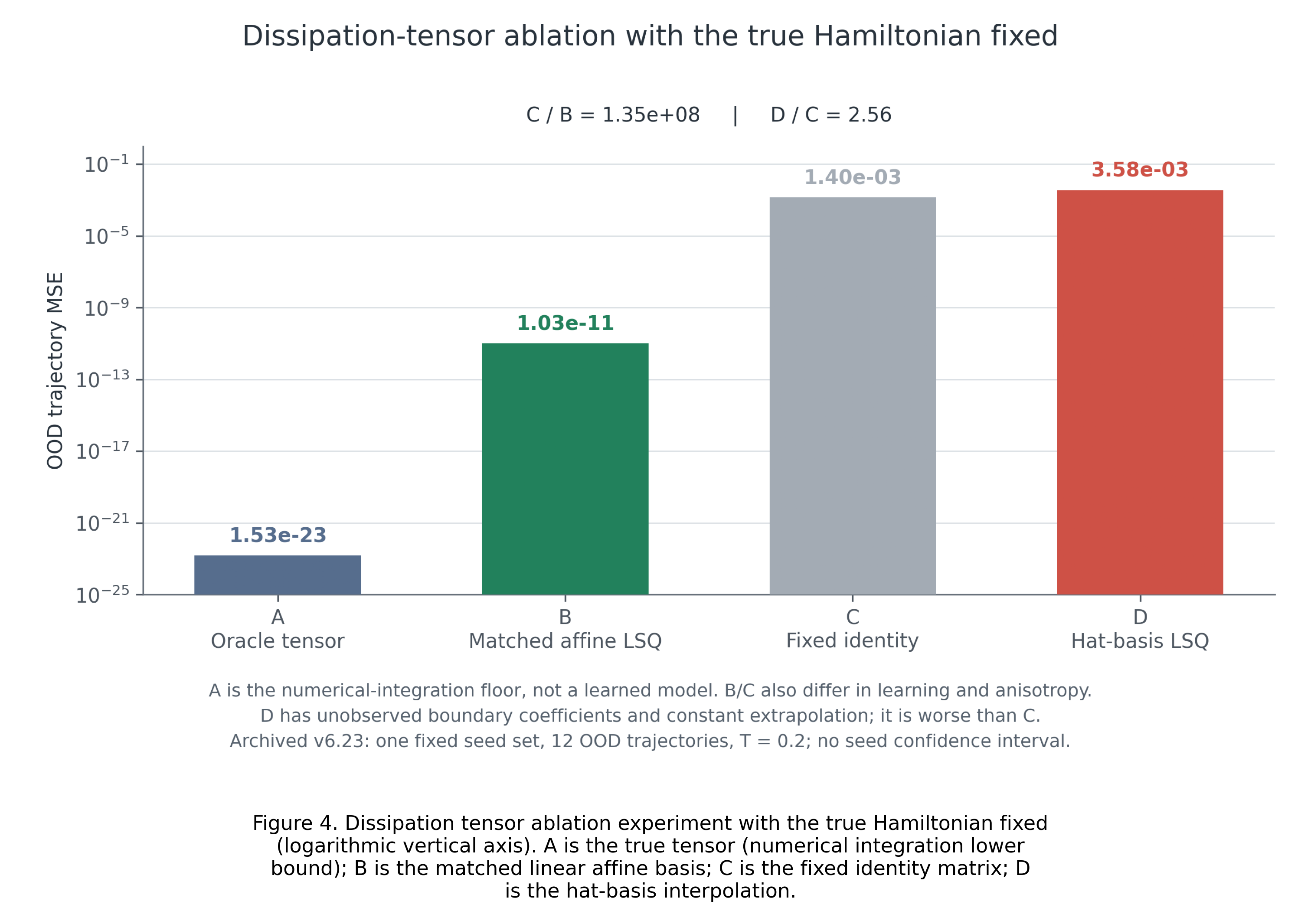}
\refstepcounter{figure}
\label{fig:case1_metric_ablation}
\end{figure}

To isolate the independent contribution of the "state-dependent metric (dissipation tensor)" itself, under the condition that the \textbf{energy structure is completely correct} (\(K\) and \(\beta\) use true values), we varied only the representation of the dissipation tensor \(G_{pp}\) and conducted four controlled comparisons (Figure \ref{fig:case1_metric_ablation}).

\textbf{Key findings}:
\begin{enumerate}
\item \textbf{The independent contribution of the state-dependent metric is enormous}: The OOD MSE of B (state-dependent) is \(1.03\times10^{-11}\), a \(1.35\times10^8\)-fold improvement over C (constant baseline, \(1.40\times10^{-3}\)).
\item \textbf{Parameterization mismatch destroys all gains}: The error of D (hat-basis interpolation, \(3.58\times10^{-3}\)) is 2.56 times worse than the constant baseline C. Because the training data does not cover the terminal nodes of the hat basis, extrapolation leads to catastrophic collapse.
\item \textbf{Perfect metric is limited by numerical precision}: The error of A (\(1.53\times10^{-23}\)) is already below double-precision floating-point noise.
\end{enumerate}
\textbf{Engineering guidance}: State-dependent dissipation is not an automatic gain. It must be paired with \textbf{identifiable parameterization} (e.g., linear basis, full-rank fitting) and \textbf{sufficient training coverage}. If using local interpolation bases, ensure the nodes cover the OOD radius.

\subsection{Sources of Advantage}
\begin{enumerate}
\item \textbf{Energy decomposition}: Potential energy is the primary source of gain (Shapley +0.80/+0.69); kernel-parameterized potential energy substantially improves over the old potential energy.
\item \textbf{Residual training}: Weak-form residuals avoid Lyapunov amplification; fine-tuning metric+energy is not interfered with by trajectory gradients.
\item \textbf{Three-stage}: Cold start allows the latent space to acquire dynamical structure; the velocity field is calibrated in this coordinate system, and finally the decoder is aligned.
\item \textbf{Metric contribution}: The state-dependent dissipation metric has a \(10^8\)-order \textbf{joint contribution} under controlled conditions; this difference simultaneously contains state dependence, anisotropy, and parameter learning, and cannot be attributed solely to "pure state dependence." The premise is that the parameterization is identifiable and training coverage is sufficient.
\end{enumerate}

\subsection{Limitations}
\begin{itemize}
\item Single system, single time domain, OOD partially overlapping.
\item The parent model condition number is approximately \(2\times10^4\)--\(3\times10^4\), exceeding the recommended threshold of 1000, yet still achieving good OOD results.
\item The historical reference target has been seen, which is exploratory interpretation.
\item Component ablation explains the difference between two frozen models and does not prove any universal causal advantage of "kernel/pullback/Fisher."
\item \textbf{Limitations of metric ablation}: The difference between B and C simultaneously contains three variables—state dependence, anisotropy, and parameter learning—and cannot independently quantify the contribution of "pure state dependence"; Group C uses a fixed \(I\) rather than the optimal constant tensor, so the B/C gap is an upper bound; the failure attribution of Group D is not fully isolated.
\end{itemize}

\section{Case 2: Target Detection Sim-to-Real Cross-Domain Fallback}

\subsection{Objective}
In industrial vision, detectors trained in simulation domains and deployed to real machines often suffer feature mismatch and localization collapse due to domain shifts in illumination, material, and background. This case validates the fallback capability of SKG in the cross-domain scenario of simulation training and real-machine testing: \textbf{no fine-tuning or domain adaptation is performed in the real-machine domain; localization performance is rescued purely by the geometric energy flow during inference.} This case is an industrial tilted wooden block Sim-to-Real cross-domain fallback (Co-DETR, simulation training, real-machine testing, unit metric + hierarchical physical contour energy).

\subsection{Applied SKG Theory}
\begin{itemize}
\item \textbf{Task invariants (\(\Delta\))}: The boundary energy is entirely driven by the image's physical contour and does not depend on network features. This is the fundamental reason the energy flow still works in cross-domain scenarios: network features fail in the real-machine domain, but the image's physical contour does not fail.
\item \textbf{Hierarchical physical contour observation (robust extraction)}: To overcome the failure of fixed color thresholds in complex backgrounds, a three-level extraction mechanism is introduced. First, a local normalized coordinate system is established based on the candidate box (core region \(r \le 0.6\), peripheral region \(r > 1.05\)). Initial contour validation is performed; if it fails, local Otsu is computed on multi-channel (grayscale/saturation/value) histograms using only the core region; if it still fails, GMM is fitted in RGB space using the core region as foreground seed and the peripheral region as background seed, and GrabCut is executed; if all stages fail, an abstention state is output and the original box is retained.
\item \textbf{Conditional target energy (\(H, S\))}:
\begin{equation}
S=\frac12\|q_{rel}-t\|^2+\frac{\lambda}{2}\|q_{rel}-q_{0,rel}\|^2,
\end{equation}
where \(t\) is the physical contour observation target state, \(q_{0,rel}\) is the initial detection box normalized state, and \(\lambda\) is the regularization weight.
\item \textbf{Angular continuity handling}: The angle of a rotated rectangle is periodic (\(\pi\) and \(-\pi\) are equivalent). Direct computation causes jumps (e.g., \(89^\circ\) and \(-89^\circ\) directly subtracted yields an error of \(178^\circ\)). A wrapped difference mechanism \(\theta_{new} = \theta_{target} + \text{wrap\_near}(\theta_{current} - \theta_{target})\) is introduced to automatically correct the error to the equivalent \(2^\circ\), ensuring smooth gradient descent.
\item \textbf{First-order dissipative flow (\(g\))}: Fixed unit metric \(g=I+10^{-6}I\), \(q \leftarrow q - \frac{1}{1+\lambda}\text{solve}(g,\nabla S)\), executed for 8 steps.
\item \textbf{Rotated IoU gating (\(\Pi\))}: Before evolution, the rotated IoU between the initial prediction box and the observation box is computed. If strictly greater than 0.9, the update is frozen and the original detection box is retained. This ensures that prediction boxes already highly aligned with the physical contour are not corrupted by contour noise.
\end{itemize}

\subsection{Parameter Settings}
\begin{table}[H]
\centering
\begin{tabular}{ll}
\toprule
Component & Configuration \\
\midrule
Detector & Co-DETR (ResNet-50), with supervised angle branch and rotated width/height branch \\
Detector training & Fine-tuned only on the simulation training set (50-epoch budget, taking epoch 48), frozen and directly used for real-machine test \\
Local coordinates & \(u, v\) normalized by detector angle rotation; core region \(r \le 0.6\), peripheral region \(r > 1.05\) \\
Contour extraction & Multi-channel local Otsu (grayscale/saturation/value) + RGB GMM-GrabCut fallback \\
State & \(q_{rel} = (\frac{c_x - c_{0x}}{L}, \frac{c_y - c_{0y}}{L}, \log\frac{w}{L}, \log\frac{h}{L}, \theta)\) \\
Energy (\(H, S\)) & Conditional target energy, angular wrapped difference \\
Flow (\(g\)) & 8-step first-order dissipation, fixed metric \(g=I+10^{-6}I\) \\
Gating (\(\Pi\)) & Pre-evolution judgment, rotated IoU > 0.9 frozen \\
Observation failure & Retain original box when all hierarchical extractions fail validation \\
\bottomrule
\end{tabular}
\end{table}

\textbf{Data}: Industrial tilted wooden block dataset. Simulation domain: 282 training images + 50 dev; real-machine domain: 32 test images. Image resolution 1280\(\times\)960, annotation format is rotated bounding box (OBB).

\subsection{Baseline Comparison Protocol}
\begin{itemize}
\item The detector is fine-tuned only on the simulation training set; the best checkpoint is selected on dev, and the weights are frozen and directly used for real-machine test.
\item No fine-tuning or domain adaptation is performed on the real-machine test.
\item Metrics: Official rotated box COCO mAP@[0.50:0.95], AP50, AP75.
\item Comparison: Same detector baseline vs. same detector + SKG energy flow, with identical candidates, classes, scores, and NMS rules.
\end{itemize}

\subsection{Results}
\begin{table}[H]
\centering
\begin{tabular}{llccc}
\toprule
Data Domain & Method & OBB mAP & AP50 & AP75 \\
\midrule
Simulation Dev & Co-DETR Baseline & 0.6469 & 0.8914 & 0.7195 \\
 & + SKG 5D Flow & \textbf{0.7687 (+12.18)} & \textbf{0.9600} & \textbf{0.8410 (+12.15)} \\
\midrule
Real-Machine Test & Co-DETR Baseline & 0.2564 & 0.6730 & 0.1736 \\
 & + SKG 5D Flow & \textbf{0.4964 (+24.00)} & \textbf{0.8432} & \textbf{0.5765 (+40.29)} \\
\bottomrule
\end{tabular}
\end{table}

\textbf{Within the simulation domain}: Baseline 0.6469. Under IoU gating protection, the SKG energy flow pulls the prediction boxes tightly to the physical contour, increasing AP75 from 0.7195 to 0.8410 (+12.15 points) and overall mAP by 12.18 points.

\textbf{Real-machine domain cross-domain collapse}: Domain shift causes catastrophic detector degradation, with mAP plummeting to 0.2564 and AP75 at only 0.1736. A large number of network prediction boxes undergo severe deformation and drift in the real-machine environment.

\textbf{Real-machine domain SKG physical fallback}: SKG energy is entirely driven by the physical contour of the real-machine image itself and does not depend on the failed network features. mAP jumps from 0.2564 to 0.4964 (+24.00 points, nearly doubling relatively), AP50 increases to 0.8432 (+17.02 points), and AP75 jumps from 0.1736 to 0.5765 (+40.29 points).

\begin{figure}[H]
\centering
\includeimage[width=0.95\textwidth]{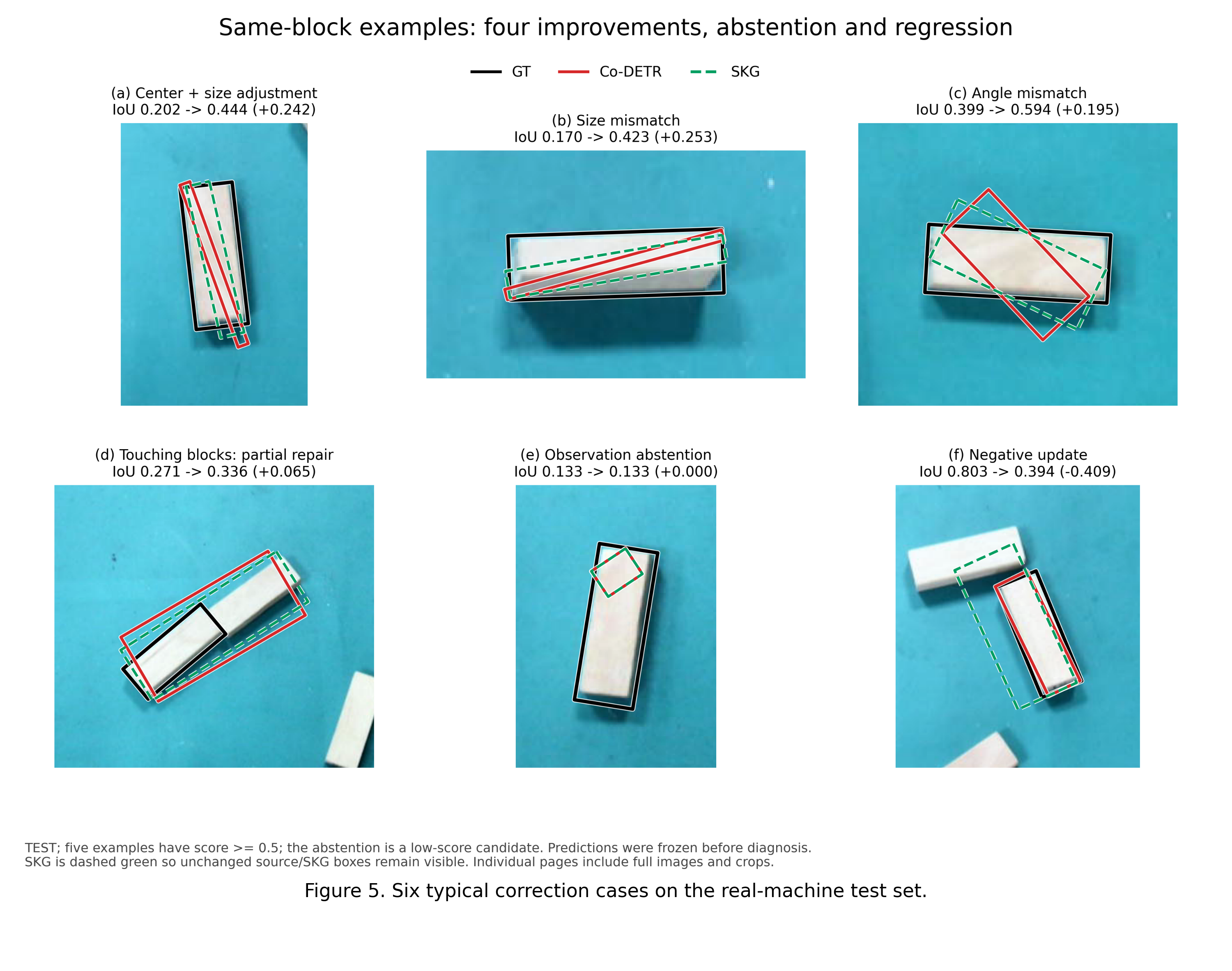}
\refstepcounter{figure}
\label{fig:case2_vis}
\end{figure}

\textbf{Qualitative visualization analysis}: Figure \ref{fig:case2_vis} shows the correction effects for 6 typical candidates (black box is GT, red box is baseline prediction, green dashed line is corrected result):
\begin{itemize}
\item (a) Center + size adjustment (IoU 0.202 \(\to\) 0.444, +0.242): The red box center is deviated and the scale mismatched; the energy flow pulls it back through the \(c_x, c_y\) and \(\log w, \log h\) dimensions.
\item (b) Size mismatch (IoU 0.170 \(\to\) 0.423, +0.253): The red box is severely enlarged; the \(\log w, \log h\) dimensions play a dominant role, shrinking it to the physical edge.
\item (c) Angle mismatch (IoU 0.399 \(\to\) 0.594, +0.195): The red box is diagonally misaligned; the 5D flow combined with wrapped difference rapidly corrects the pose.
\item (d) Touching blocks: Partial repair (IoU 0.271 \(\to\) 0.336, +0.065): Contour extraction is disturbed by adjacent objects; SKG performs a certain degree of geometric correction.
\item (e) Observation abstention (IoU 0.133 \(\to\) 0.133, +0.000): Due to connected components touching the boundary or graph cut failure, the system triggers abstention and retains the original box.
\item (f) Negative update (IoU 0.803 \(\to\) 0.394, -0.409): In extreme cases, if the background chromaticity is extremely similar to the foreground causing incorrect contour extraction, the energy flow will pull the originally correct box off.
\end{itemize}

\textbf{Statistical note}: This case is a single-run experiment with a small real-machine test sample size (32 images); no statistical significance test is performed; the mAP improvement is a point estimate.

\subsection{Mechanism Validation}
\begin{figure}[H]
\centering
\includeimage[width=0.9\textwidth]{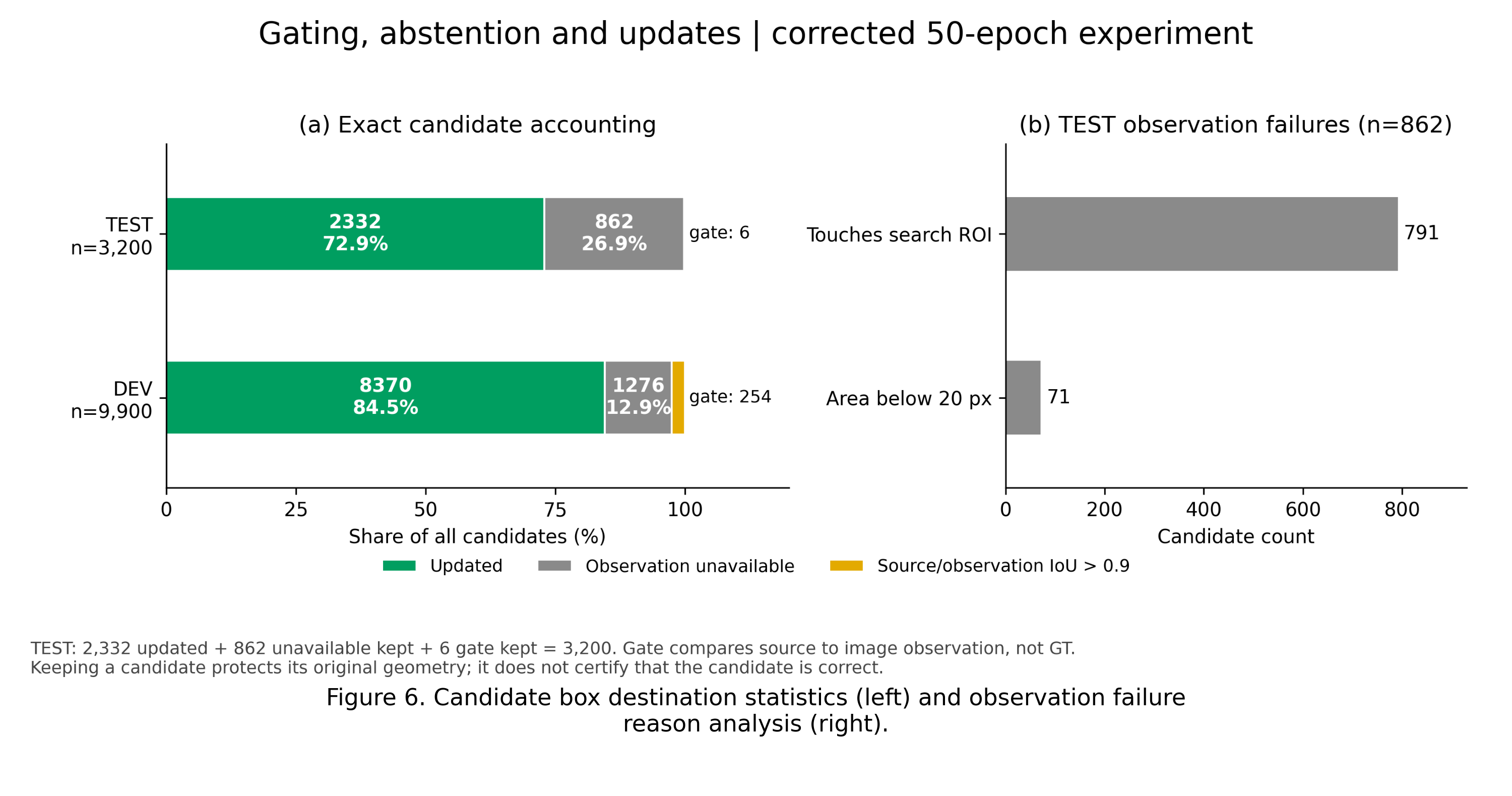}
\refstepcounter{figure}
\label{fig:case2_stats}
\end{figure}

\textbf{Candidate destination and gating statistics}:
The total number of real-machine test candidates is 3,200 (including all classes/all scores). The destinations of these 3,200 real-machine candidates are: 2,332 (72.9\%) successfully evolved and corrected, 862 (26.9\%) retained original boxes due to observation failure, and 6 frozen by gating (IoU>0.9). The main reasons for observation failure are connected components touching the search ROI boundary (791) or area too small (71) (Figure \ref{fig:case2_stats}).

\begin{figure}[H]
\centering
\includeimage[width=0.8\textwidth]{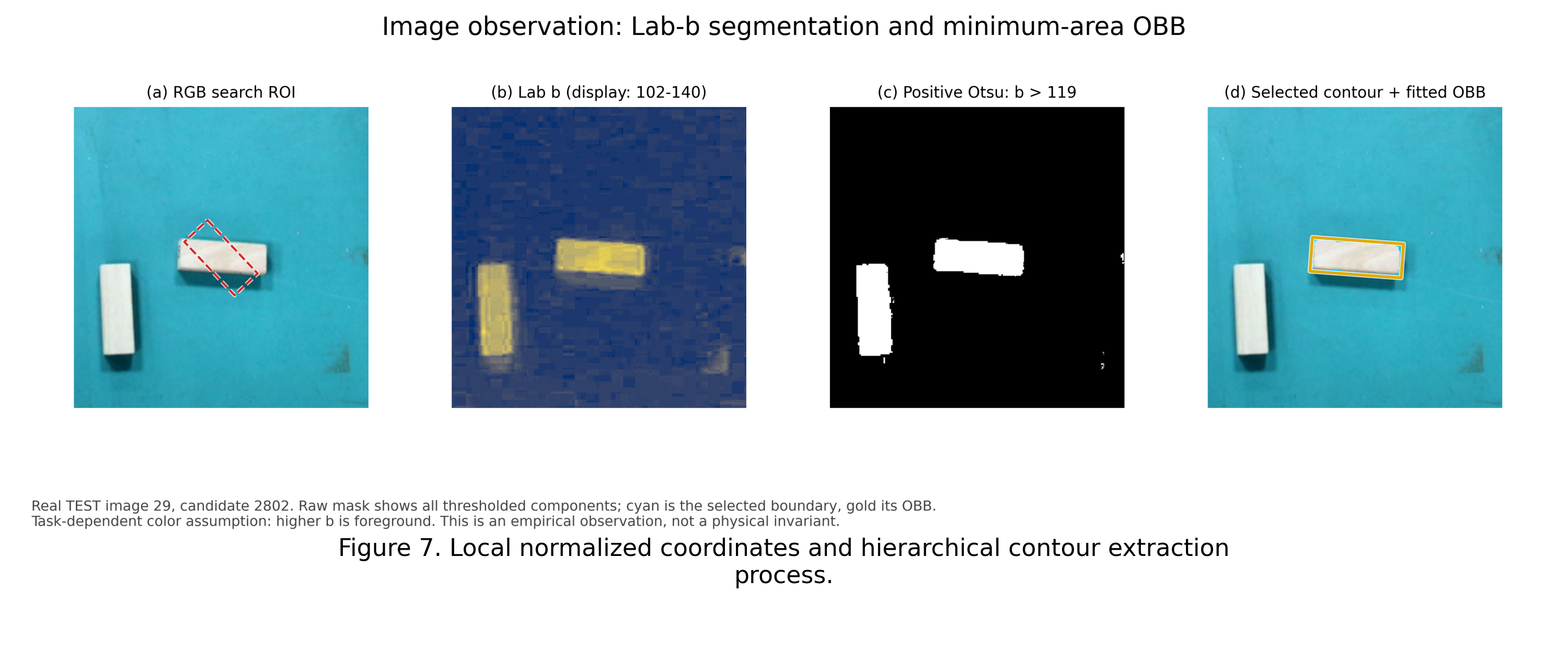}
\refstepcounter{figure}
\label{fig:case2_mask}
\end{figure}

\begin{figure}[H]
\centering
\includeimage[width=0.8\textwidth]{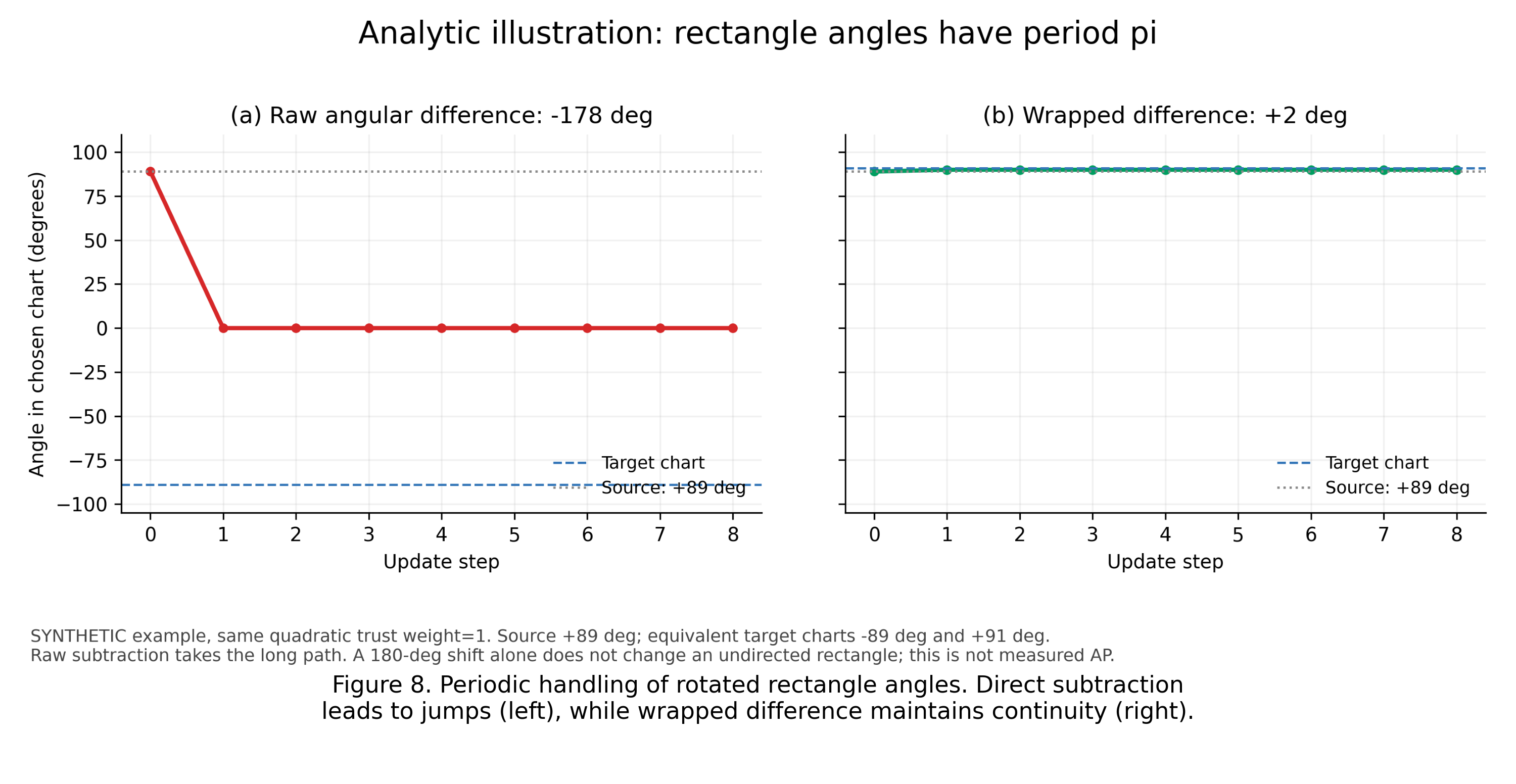}
\refstepcounter{figure}
\label{fig:case2_angle}
\end{figure}

\textbf{Hierarchical physical contour extraction and angular wrapping}:
Figure \ref{fig:case2_mask} shows the local coordinate establishment and hierarchical contour extraction process. When multi-channel Otsu fails, GMM-GrabCut using the core region as foreground seed and the peripheral region as background seed can effectively salvage extraction. Figure \ref{fig:case2_angle} shows the necessity of angular wrapping: if directly subtracted (left figure), the angle jumps from +89° to -91° (difference of -178°); using wrapped difference (right figure), the angle difference remains at +2°, avoiding gradient explosion and reverse rotation.

\begin{figure}[H]
\centering
\includeimage[width=0.9\textwidth]{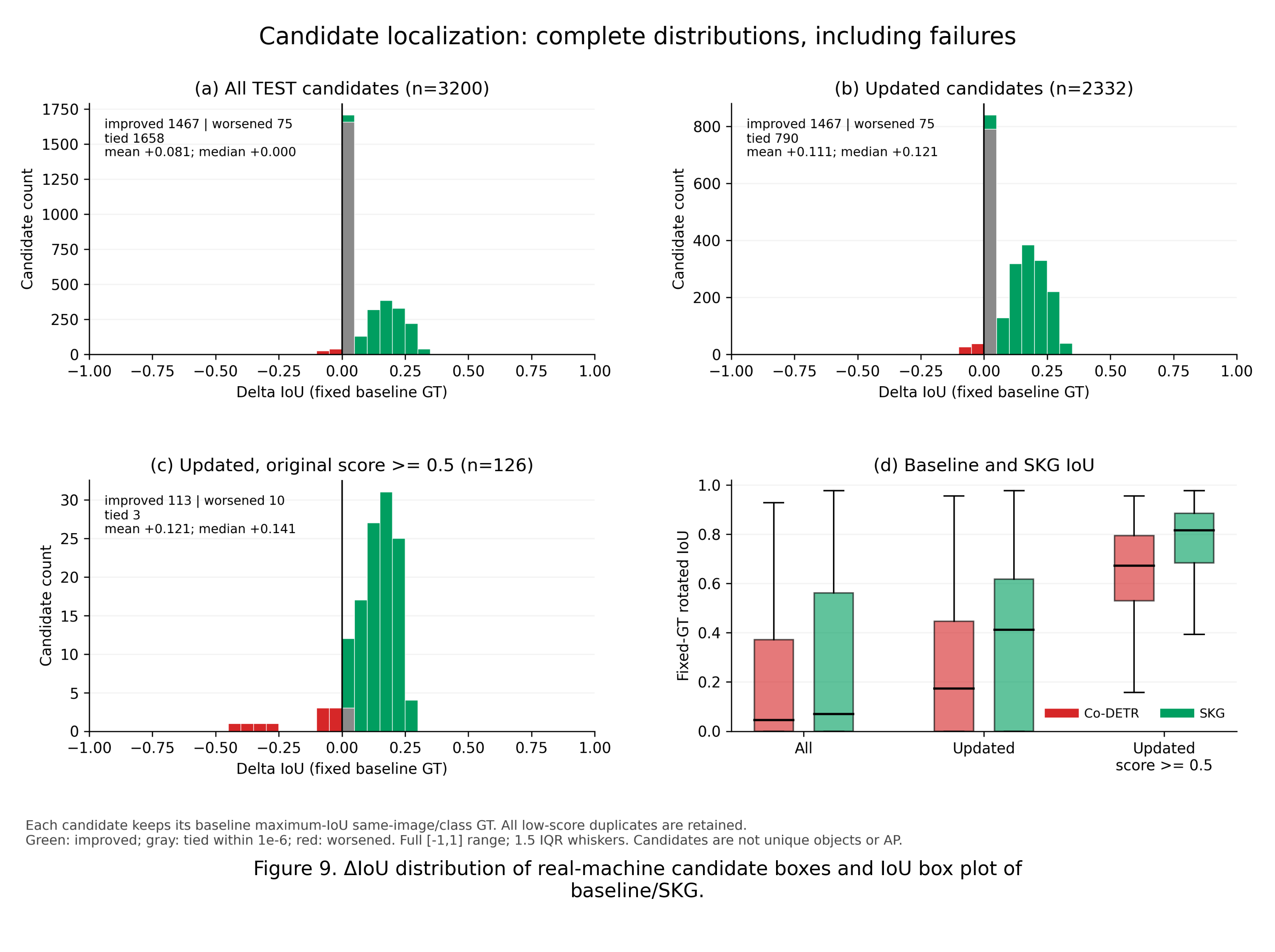}
\refstepcounter{figure}
\label{fig:case2_iou}
\end{figure}

\textbf{IoU improvement distribution}:
Figure \ref{fig:case2_iou} shows the histogram and box plot of candidate box IoU improvement (\(\Delta\)IoU). Among all 3,200 real-machine candidates, 1,467 improved, 75 worsened, and 1,658 remained unchanged (mostly abstention boxes from observation failure), with an average improvement of +0.081. Among the 2,332 successfully updated candidates, the average improvement is +0.111, and the median improvement is +0.121. For the 126 high-confidence candidates with original confidence \(\ge 0.5\), the average improvement is +0.121, and the worsening ratio is extremely low (only 10).

\subsection{Sources of Advantage}
\begin{enumerate}
\item \textbf{Physical geometric prior does not depend on network features}: SKG energy is entirely driven by hierarchical physical contours. When the detector collapses in the real-machine domain due to feature mismatch, the energy flow can still work. This is the fundamental reason for cross-domain fallback.
\item \textbf{Robustness of hierarchical extraction}: Through the four-level mechanism of "local validation \(\to\) multi-channel Otsu \(\to\) GMM-GrabCut graph cut \(\to\) abstention," the success rate of contour extraction under complex backgrounds, weak contrast, and uneven illumination is greatly improved.
\item \textbf{Angular continuity wrapped difference}: Completely solves the jump problem of the energy flow in angle dimension computation, preventing the correction process from becoming "reverse optimization."
\item \textbf{Huge gains in high-precision intervals}: The real-machine AP75 improvement of 40.29 points shows that the physical fallback has the most significant rescue effect on strict localization (high IoU).
\end{enumerate}

\subsection{Limitations}
\begin{itemize}
\item Real-machine test has only 32 images, limited sample size.
\item The degree of domain shift between simulation dev and real-machine test is not quantified.
\item Although hierarchical contour extraction is powerful, it still relies on a certain degree of foreground/background separability. In extreme cases such as complete color similarity between target and background, or severe occlusion, incorrect contours may be extracted, leading to very few negative updates (as shown in Figure \ref{fig:case2_vis}(f), where IoU 0.803 degrades to 0.394).
\item This case uses the unit metric \(g=I+10^{-6}I\). The value of the pulled-back metric can only potentially emerge under the Fisher objective metric; this case does not implement Fisher.
\end{itemize}

\section{Case 3: Sequence Modeling sine\(\to\)square}

\subsection{Objective}
To validate the OOD generalization of SKG under distribution shift. Train on sine, test on square; both share frequency and amplitude ranges but have different waveform structures.

\subsection{Applied SKG Theory}
\begin{itemize}
\item \textbf{Energy decomposition (\(H, S\))}: The energy is a sum of four terms, each contributing a different geometric shape in the 16-dimensional latent space. Their superposition forms the complete "energy landscape":
\begin{equation}
E(z) = \underbrace{\tfrac12\lambda\|z\|^2}_{\text{quad}} + \underbrace{\beta\,\mathrm{relu}(\|z\|^2-r^2)^2}_{\text{bound}} + \underbrace{\mathrm{MLP}(|z|)}_{\text{mlp}} + \underbrace{w\,\mathrm{SpecMLP}(|\mathrm{rfft}(Wz)|)}_{\text{spec}}.
\end{equation}
\begin{itemize}
\item \textbf{quad} \(\tfrac12\lambda\|z\|^2\): A parabolic bowl centered at the origin. It serves as the \textbf{global bowl bottom}, ensuring \(z\) is bounded and energy monotonically increases far away.
\item \textbf{bound} \(\beta\,\mathrm{relu}(\|z\|^2-r^2)^2\): A quartic wall outside radius \(r\). It provides a \textbf{hard boundary}, where energy steeply increases outside the sphere, confining the evolution within the ball.
\item \textbf{mlp} \(\mathrm{MLP}(|z|)\): Even-symmetric arbitrary undulations. It creates \textbf{local minima, saddle points, and ridges}—the main "terrain details".
\item \textbf{spec} \(w\,\mathrm{SpecMLP}(|\mathrm{rfft}(Wz)|)\): Undulations along the spectral direction. It makes the landscape sensitive to the spectral structure of the signal and is the \textbf{only term that can possibly "recognize the square wave"} during OOD.
\end{itemize}
\item \textbf{Overall landscape shape}: Three layers can be seen: the \textbf{outer layer} (quad + bound) forms a finite-depth bowl (conservative, isotropic); the \textbf{middle layer} (mlp) carves multiple local minima at the bottom (task-relevant, shaped by trajectory training); the \textbf{inner layer} (spec) adds undulations along the spectral direction (anisotropic, sensitive to specific directions).
\item \textbf{Evolution in the landscape}: Inference is a 10-step gradient descent \(z_{t+1}=z_t-\eta(z_t)\,G^{-1}(z_t)\nabla E(z_t)\). The starting point \(z_0=\mathcal E(x)\) falls inside the bowl, and each step moves along \(-\nabla E\), with the metric \(G^{-1}\) converting gradients into velocity. Backtracking search ensures energy does not increase at each step. After 10 steps, it stops near a local minimum, where the decoder reads \(\hat y\).
\item \textbf{Three-stage training}: Trajectory cold start \(\to\) weak-form residual fine-tuning of metric+energy \(\to\) trajectory alignment of decoder.
\item \textbf{Weak-form residual training}: Simpson local integration.
\end{itemize}

\subsection{Parameter Settings}
\begin{table}[H]
\centering
\begin{tabular}{lll}
\toprule
Component & Form & Parameter Count \\
\midrule
State & Encoder MLP(24\(\to\)64\(\to\)16) & \(\sim\)1.7K \\
Metric (\(g\)) & \(G(z)=\operatorname{diag}(d)+UU^\top\), rank 4 & \(\sim\)1.8K \\
Energy (\(H, S\)) & quad + bound + mlp + spec & \(\sim\)5.5K \\
Step size & \(\eta(z)=\exp(\log\eta_0+\mathrm{MLP}(z))\) & \(\sim\)500 \\
Evolution & 10-step explicit evolution + 3 backtracking & --- \\
Decoder & Linear layer \(16\to1\) & 17 \\
\midrule
\textbf{Total} & & \textbf{9,489} \\
\bottomrule
\end{tabular}
\end{table}

\textbf{Three stages}:
\begin{table}[H]
\centering
\begin{tabular}{clcl}
\toprule
Stage & Training Object & Loss & Epoch \\
\midrule
A & encoder + metric + energy + decoder + \(\eta\) & Trajectory loss & 100 \\
B & metric + energy + \(\eta\) & Weak-form residual (Simpson) & 100 \\
C & decoder & Trajectory loss & 50 \\
\bottomrule
\end{tabular}
\end{table}

\textbf{Data}: Window 24, frequency 0.1, noise 0.05. Train 3000 single windows + 1000 trajectories, validation 500, ID test 500, OOD test 500.

\subsection{Baseline Comparison Protocol}
\begin{itemize}
\item Same data split (3000 single windows training, 500 validation, 500 ID test, 500 OOD test).
\item Comparison: 1-layer Transformer, d\_model 32, 11,617 parameters.
\item Metrics: ID MSE, OOD MSE, OOD/ID.
\item Paired t-test, 5 seeds.
\end{itemize}

\subsection{Results and Inference Budget Analysis}
\textbf{Quantitative comparison (mean of 5 seeds)}:
\begin{table}[H]
\centering
\begin{tabular}{lcc}
\toprule
Metric & SKG & Transformer \\
\midrule
ID MSE & 0.003023 \(\pm\) 0.000223 & 0.003099 \(\pm\) 0.000296 \\
OOD MSE & \textbf{0.029577 \(\pm\) 0.008228} & 0.064096 \(\pm\) 0.018351 \\
OOD/ID & \textbf{9.953 \(\pm\) 3.251} & 21.036 \(\pm\) 6.893 \\
\bottomrule
\end{tabular}
\end{table}

\begin{figure}[H]
\centering
\includeimage[width=0.8\textwidth]{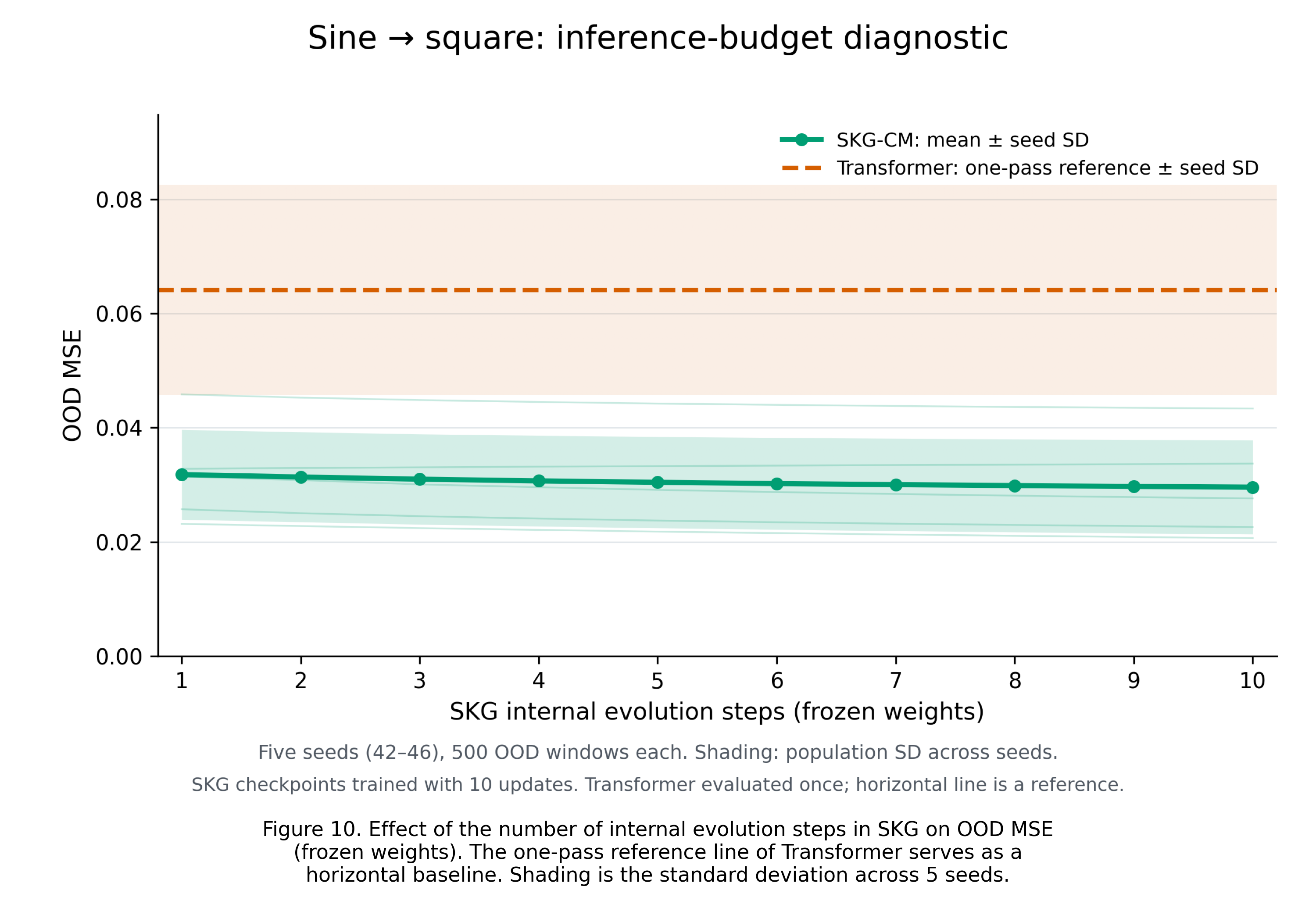}
\refstepcounter{figure}
\label{fig:case3_budget}
\end{figure}

\textbf{Inference budget diagnostic (internal evolution steps with frozen weights)}:
Under the premise of freezing network weights, we varied the number of internal evolution steps during SKG inference (1 to 10) and observed its impact on OOD MSE (Figure \ref{fig:case3_budget}). With only \textbf{1-step inference}, OOD MSE is \textbf{0.031776}; with the preset \textbf{10-step inference}, OOD MSE decreases to \textbf{0.029577}. This indicates that the geometric evolution mechanism during inference itself brings additional generalization gain, and this gain smoothly accumulates with increasing evolution steps.

\begin{figure}[H]
\centering
\includeimage[width=0.8\textwidth]{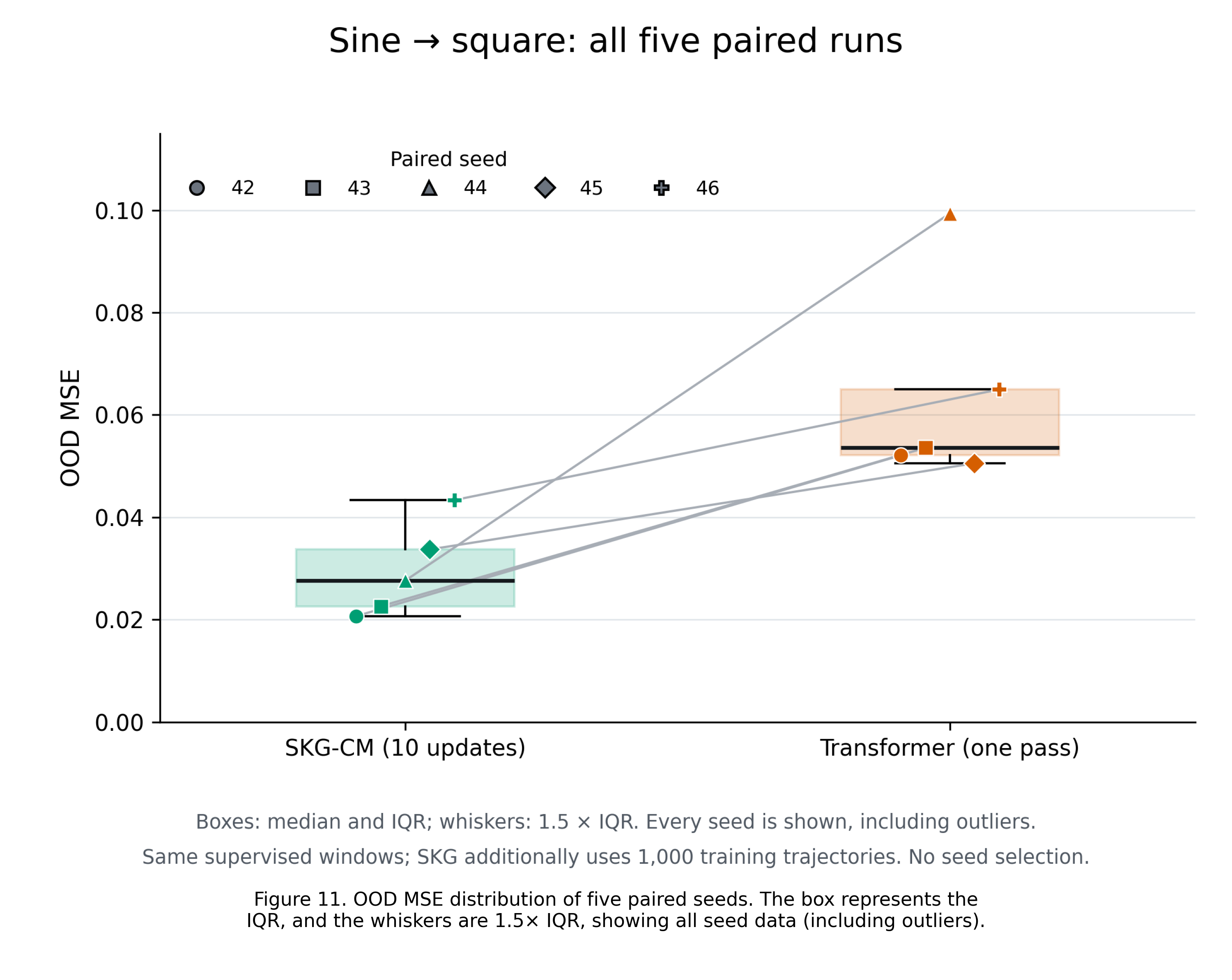}
\refstepcounter{figure}
\label{fig:case3_seeds}
\end{figure}

\textbf{Multi-seed paired distribution}:
Figure \ref{fig:case3_seeds} shows the OOD MSE distribution and box plot of 5 paired seeds (42--46). SKG-CM (green box) outperforms Transformer (orange box) on all 5 seeds, and the paired lines (gray lines) consistently slope upward to the right, indicating that Transformer's error is significantly higher on every seed.

\begin{figure}[H]
\centering
\includeimage[width=0.8\textwidth]{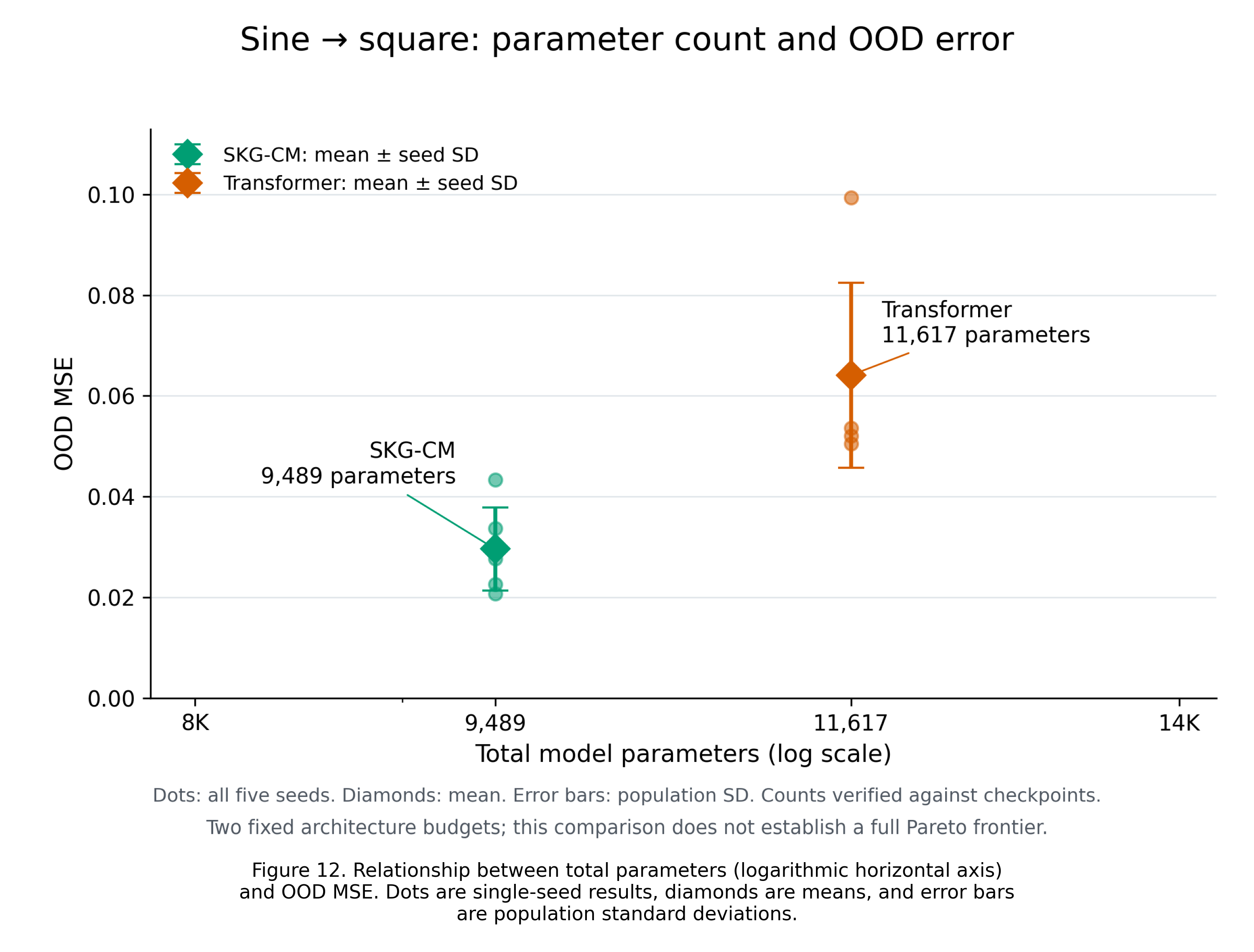}
\refstepcounter{figure}
\label{fig:case3_params}
\end{figure}

\textbf{Parameter efficiency comparison}:
Figure \ref{fig:case3_params} shows the relationship between model parameter count and OOD MSE. SKG-CM achieves an average OOD MSE of \textbf{0.0296} with \textbf{9,489} parameters, while Transformer uses \textbf{11,617} parameters with an average OOD MSE of \textbf{0.0641}. SKG reduces OOD error by 54\% while using 18.3\% fewer parameters.

\textbf{Paired t-test}: ID \(p=0.254\); OOD \(p=0.024\), mean difference \(-0.034520\).

\subsection{Sources of Advantage}
\begin{enumerate}
\item \textbf{Evolution structure (multi-step buffering)}: Under fixed weights, increasing inference evolution steps from 1 to 10 reduces OOD MSE from 0.031776 to 0.029577. Each evolution step is constrained by the metric/energy landscape, and the shift in the latent space during OOD is buffered by the dynamical structure.
\item \textbf{Energy landscape prior for the square wave}: The spectral prior (spec term) is the \textbf{only explicit component sensitive to the spectrum}, detecting the odd harmonics of the square wave through the spectral MLP and altering the local shape of the landscape near the square wave's starting point. This is likely one of the reasons SKG does not degrade on OOD—the landscape itself "knows" the waveform has changed.
\item \textbf{Alignment loss}: Landscape alignment and state direction alignment provide additional gradient signals.
\item \textbf{Limited marginal contribution of metric}: The OOD of the position-dependent metric and the Identity metric are almost identical (0.0244 vs 0.0243). Under the condition that the window covers only 2.4\% of the period, the OOD advantage mainly comes from evolution structure + energy landscape prior + alignment loss.
\item \textbf{High parameter efficiency}: With fewer parameters (9.5K vs 11.6K), it achieves lower OOD error and higher stability on all 5 seeds.
\end{enumerate}

\subsection{Limitations}
\begin{itemize}
\item \textbf{Energy minima do not correspond to the task ground truth}: Validation shows that at the true state \(z^*\), the gradient \(\|\nabla E(z^*)\|\approx317\) and the Hessian has negative eigenvalues; \(z^*\) is \textbf{not} a local minimum. Therefore, the decoder learns to map from the landscape's local minima to \(y\), not from the true value to \(y\). The landscape is self-consistent (evolution reaches a minimum, energy decreases), but the semantic meaning of the minima is \textbf{learned}, not physically given.
\item Paired t-test for OOD \(p=0.024\).
\item SKG has 18.3\% fewer parameters than Transformer but 1000 more trajectories in training data.
\item The OOD MSE of seeds 45 and 46 are 0.0337 and 0.0433 respectively, and seed 42 is 0.0207 (see Figure \ref{fig:case3_seeds}).
\item Figure \ref{fig:case3_params} is only a comparison of two fixed architecture budgets and does not represent the Pareto frontier at arbitrary parameter scales.
\end{itemize}

\section{Case 4: Molecular Dynamics MD17}

\subsection{Objective}
To validate the OOD generalization of SKG on second-order Newtonian dynamics and compare two types of parameterization: potential energy residual and trajectory residual. Given two consecutive frames of atomic positions, predict future 10/50/100 step trajectories.

\subsection{Applied SKG Theory}
\begin{itemize}
\item \textbf{Energy decomposition (\(H, S\))}: \(V=V_{\text{MACE}}+\Delta V_\theta\). \(V_{\text{MACE}}\) is the frozen known physics \cite{MACEOFF}, and \(\Delta V_\theta\) is the residual potential energy to be learned.
\item \textbf{Learning decoupling}: MACE is frozen (millions of parameters), and the only training object is \(\Delta V_\theta\) (approximately 3K parameters).
\item \textbf{Zero initialization}: The last layer weights are zeroed; before training, \(\Delta V_\theta\equiv0\), strictly degenerating to the MACE-Cart baseline.
\item \textbf{Strong-form residual training}: The training objective is force residual MSE.
\end{itemize}

\subsection{Parameter Settings}
\textbf{Potential energy residual}:
\begin{table}[H]
\centering
\begin{tabular}{ll}
\toprule
Component & Configuration \\
\midrule
Baseline & MACE-OFF23 medium (frozen) \\
Residual network & Conservative Residual EGNN \\
Hidden dimension & 64 \\
Layers & 3 \\
RBF count & 32 \\
Cutoff radius & 5.0 \AA \\
Trainable parameters & \(\sim\)3K \\
Zero initialization & Yes \\
Training objective & Force residual MSE \\
Optimizer & AdamW, lr \(10^{-3}\), cosine annealing to \(5\times10^{-5}\) \\
Batch size & 8 \\
Epochs & 100 \\
\bottomrule
\end{tabular}
\end{table}

\textbf{Trajectory residual proxies}:
\begin{table}[H]
\centering
\begin{tabular}{llcl}
\toprule
Method & Structure & Parameters & Training Objective \\
\midrule
ESTAG-res & Temporal graph network + equivariant message passing & 478,622 & \(\|\Delta R_\theta-(R_{\text{true}}-R_{\text{MACE}})\|^2\) \\
ATOM-res & Global neural operator & \(\sim\)1M & Same as above \\
GSE-res & Velocity field integration & --- & Same as above \\
\bottomrule
\end{tabular}
\end{table}
All trajectory residuals share: 500 training frames, 100 epochs, AdamW, lr=\(10^{-3}\), batch=4, identical MACE-Cart frozen baseline.

\textbf{Data}: MD17 Aspirin, NVT 500K, dt=0.5fs. Train 500, validation 100, ID 100, OOD 100.

\subsection{Baseline Comparison Protocol}
\begin{itemize}
\item All methods use the same split files, the same anchor frames, and the same 50/100 frame generation history.
\item Input: Two frames of historical coordinates + atomic numbers.
\item Output: Future \(n\)-step trajectory.
\item Metrics: ADE (per-atom average displacement error), FDE (per-atom final displacement error).
\item Checkpoint is selected solely by validation set ADE@10.
\end{itemize}

\textbf{External baselines}:
\begin{table}[H]
\centering
\begin{tabular}{lcl}
\toprule
Baseline & Parameters & Source \\
\midrule
ESTAG \cite{ESTAG} & 7,799 & NeurIPS 2023 \\
GSE-Flow \cite{GSEFlow} & 25,208 & ICML 2026 \\
GF-NODE \cite{GFNODE} & 531,986 & TMLR 2025 \\
ATOM \cite{ATOM} & 3,557,830 & ICLR 2026 \\
MACE-Cart & 0 (frozen) & Reference baseline \\
\bottomrule
\end{tabular}
\end{table}

\subsection{Results}
\textbf{External baseline comparison (ID, ADE unit \AA)}:
\begin{table}[H]
\centering
\begin{tabular}{lccccc}
\toprule
Model & Parameters & @10 & @50 & @100 \\
\midrule
\textbf{SKG Potential Residual} & \textbf{\(\sim\)3K} & \textbf{0.003853} & \textbf{0.017989} & \textbf{0.028546} \\
ESTAG & 7,799 & 0.006718 & 0.065849 & 0.146704 \\
MACE-Cart & 0 & 0.006710 & --- & --- \\
GF-NODE & 531,986 & 0.037661 & 0.145711 & 0.219097 \\
GSE-Flow & 25,208 & 0.037954 & 0.223124 & 2.761314 \\
ATOM & 3,557,830 & 0.068298 & 0.718328 & 1.152361 \\
\bottomrule
\end{tabular}
\end{table}

\textbf{External baseline comparison (OOD, ADE unit \AA)}:
\begin{table}[H]
\centering
\begin{tabular}{lccc}
\toprule
Model & @10 & @50 & @100 \\
\midrule
\textbf{SKG Potential Residual} & \textbf{0.004144} & \textbf{0.018740} & \textbf{0.031575} \\
ESTAG & 0.006977 & 0.063881 & 0.141094 \\
\bottomrule
\end{tabular}
\end{table}

\textbf{Improvement relative to the second-best baseline (ESTAG)}: ID @10 \(-42.6\%\), ID @100 \(-80.5\%\), OOD @10 \(-40.6\%\), OOD @100 \(-77.6\%\).

\begin{figure}[H]
\centering
\includeimage[width=0.95\textwidth]{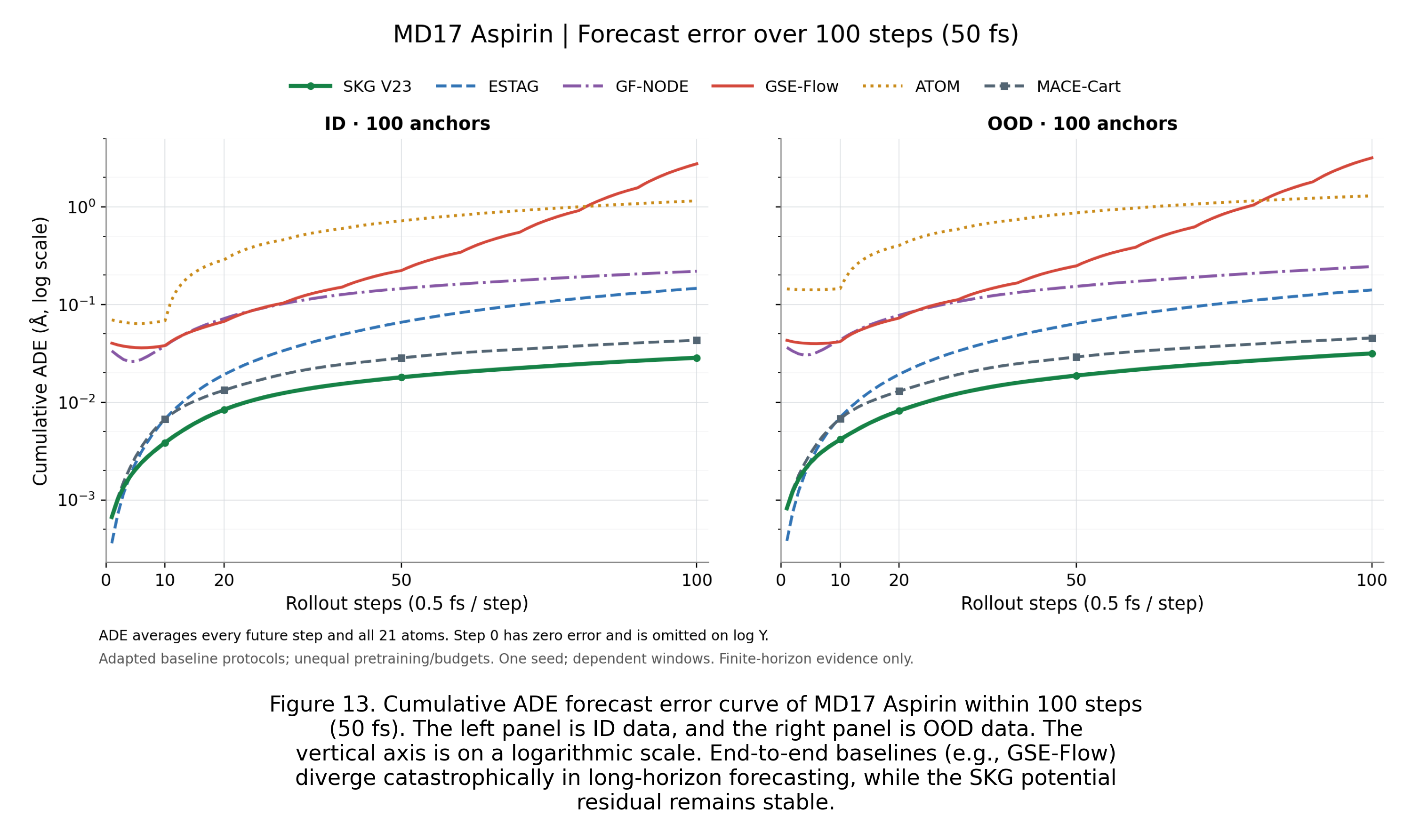}
\refstepcounter{figure}
\label{fig:case4_curve}
\end{figure}

\textbf{Long-horizon error accumulation trend}:
To intuitively demonstrate the stability of different methods in long-horizon prediction, we plotted the relationship between prediction steps and cumulative ADE error (Figure \ref{fig:case4_curve}). Under both ID and OOD splits, the errors of end-to-end methods (GSE-Flow, ATOM) exhibit exponential explosive growth with prediction steps. In contrast, the error curve of the SKG potential residual (SKG V23) maintains an extremely gentle linear growth within 100 steps, and at @100 steps its error is two orders of magnitude lower than GSE-Flow. This indicates that the physical constraint based on local force residuals successfully suppresses error accumulation in autoregressive prediction.

\begin{figure}[H]
\centering
\includeimage[width=0.9\textwidth]{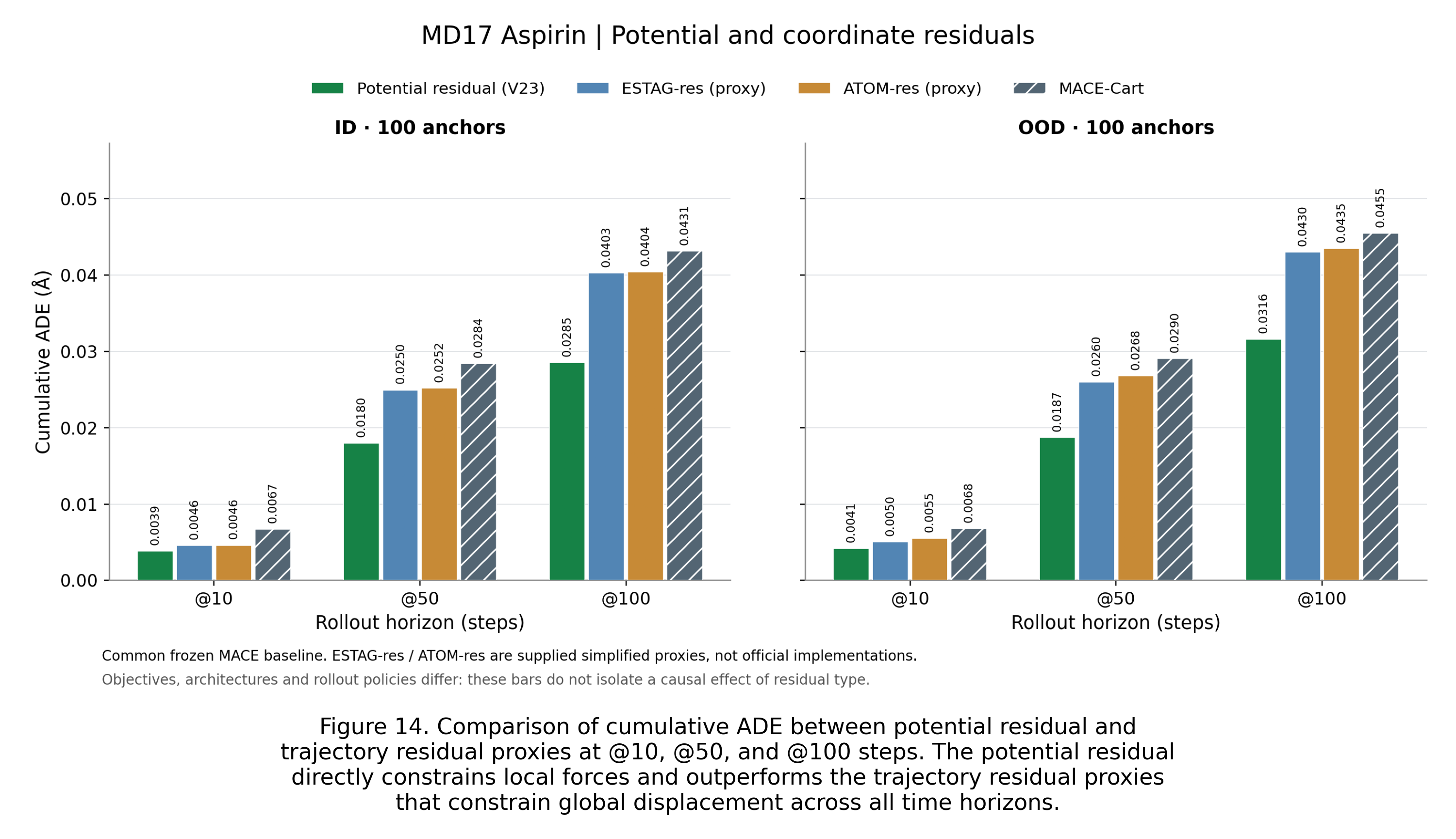}
\refstepcounter{figure}
\label{fig:case4_bar}
\end{figure}

\textbf{Potential residual vs. trajectory residual proxies}:
To isolate the impact of "residual type" on prediction, we constructed trajectory residual proxy networks based on the same frozen MACE-Cart baseline and compared them with the potential residual (Figure \ref{fig:case4_bar}).

\textbf{Note}: ESTAG-res / ATOM-res in the table are trajectory residual proxies constructed in this paper, not the official implementations of ESTAG / ATOM; the parameter count differences arise from the proxy structures and cannot be directly compared with official baselines.

\textbf{Potential residual vs. trajectory residual (ID)}:
\begin{table}[H]
\centering
\begin{tabular}{lcccccc}
\toprule
Method & Parameters & @10 & @50 & @100 & Relative to Potential @10 & @100 \\
\midrule
\textbf{Potential Residual} & \textbf{\(\sim\)3K} & \textbf{0.003853} & \textbf{0.017989} & \textbf{0.028546} & --- & --- \\
ESTAG-res & 478,622 & 0.004567 & 0.024961 & 0.040293 & +18.5\% & +41.2\% \\
ATOM-res & \(\sim\)1M & 0.004591 & 0.025167 & 0.040385 & +19.2\% & +41.5\% \\
MACE-Cart & 0 & 0.006706 & 0.028402 & 0.043124 & +74.0\% & +51.0\% \\
\bottomrule
\end{tabular}
\end{table}

\textbf{Potential residual vs. trajectory residual (OOD)}:
\begin{table}[H]
\centering
\begin{tabular}{lccc}
\toprule
Method & @10 & @50 & @100 \\
\midrule
\textbf{Potential Residual} & \textbf{0.004144} & \textbf{0.018740} & \textbf{0.031575} \\
ESTAG-res & 0.005046 & 0.025979 & 0.042990 \\
ATOM-res & 0.005502 & 0.026796 & 0.043456 \\
MACE-Cart & 0.006763 & 0.029043 & 0.045503 \\
\bottomrule
\end{tabular}
\end{table}

\textbf{Advantage expands with rollout length}: @10 \(-15.6\%\), @50 \(-27.9\%\), @100 \(-29.2\%\) (ID, versus ESTAG-res).

\begin{figure}[H]
\centering
\includeimage[width=0.95\textwidth]{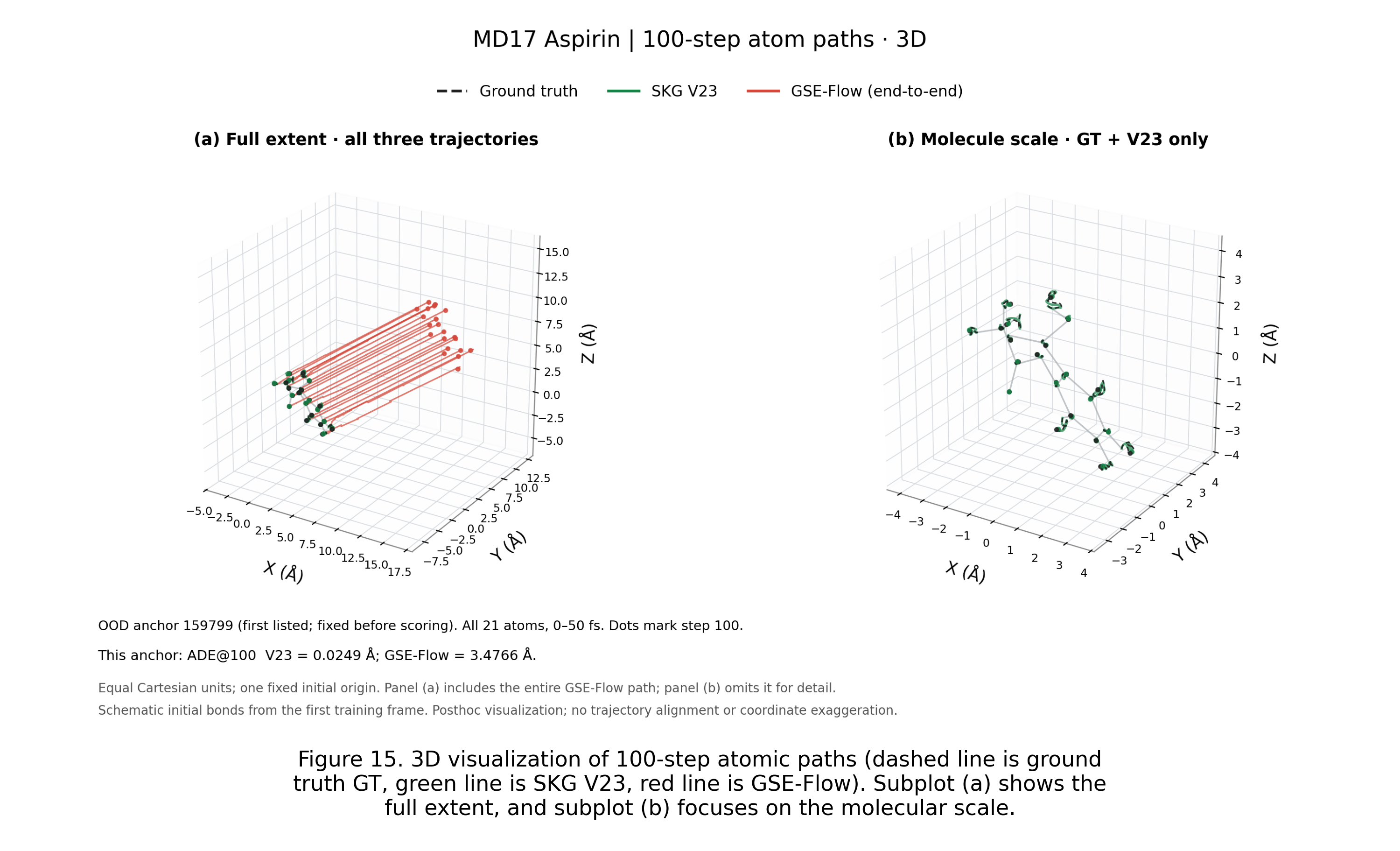}
\refstepcounter{figure}
\label{fig:case4_3d}
\end{figure}

\begin{figure}[H]
\centering
\includeimage[width=0.95\textwidth]{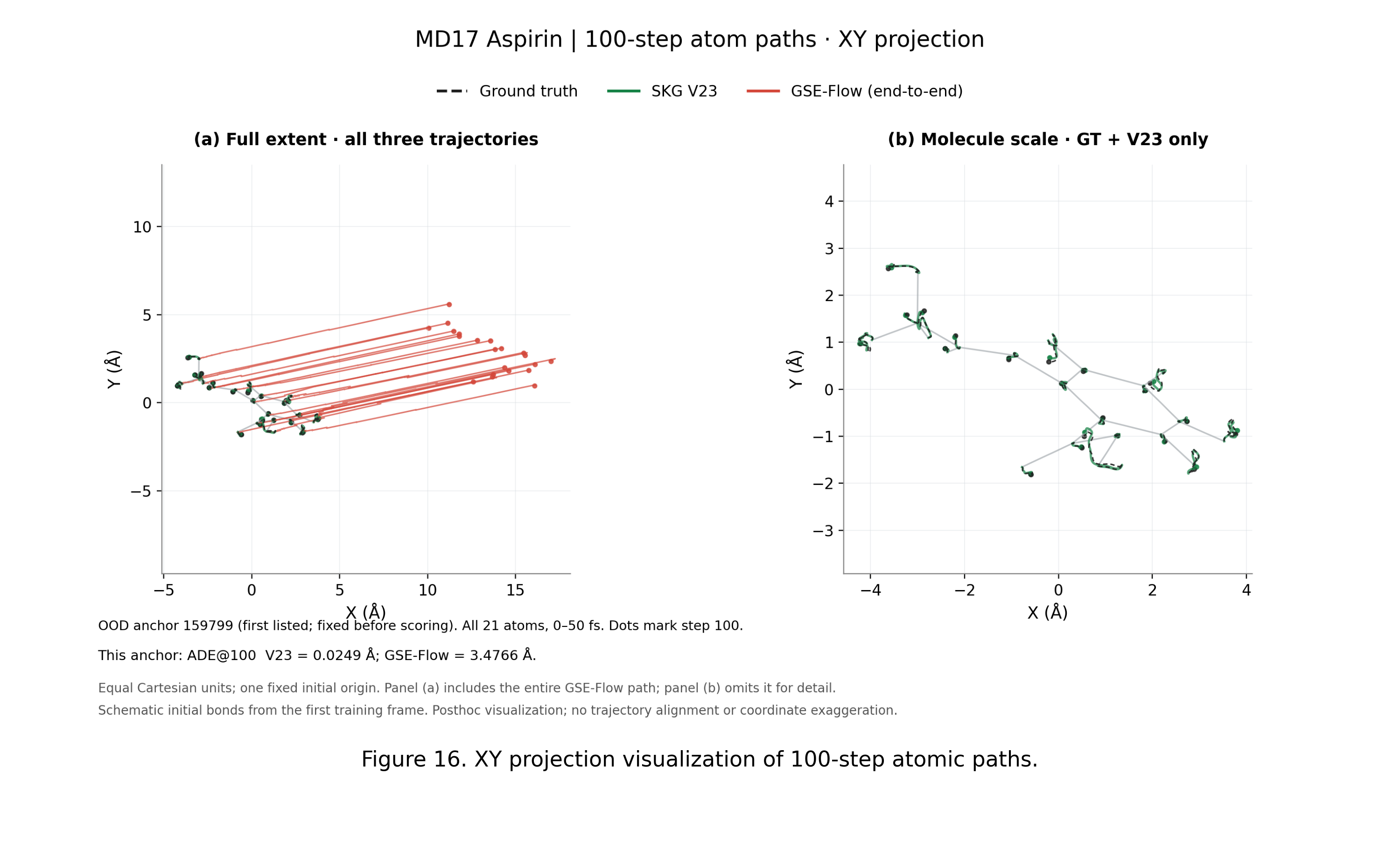}
\refstepcounter{figure}
\label{fig:case4_xy}
\end{figure}

\textbf{3D trajectory visualization}:
Figures \ref{fig:case4_3d} and \ref{fig:case4_xy} show the atomic motion trajectories of a typical anchor (anchor 159799) on the OOD test set over 100 steps.

It is clearly visible from the visualization results that, due to the lack of physical law constraints, the end-to-end method (GSE-Flow) rapidly collapses in long-horizon autoregressive prediction, with predicted atomic positions completely flying out of the true molecular motion range. In contrast, the SKG potential residual, through the base physical constraints provided by the frozen MACE and paired with small residual corrections, makes the generated trajectory still highly coincident with the true physical trajectory after 100 steps (50 fs).

\subsection{Sources of Advantage}
\begin{enumerate}
\item \textbf{Energy decomposition}: \(V=V_{\text{MACE}}+\Delta V_\theta\), known physics frozen, only residual learned.
\item \textbf{Learning decoupling}: The only learning object is \(\Delta V_\theta\) (approximately 3K parameters), which is 1/160 of the parameter count of the trajectory residual proxies.
\item \textbf{Zero initialization}: Before training, it strictly degenerates to MACE-Cart, and the gain clearly comes from residual correction.
\item \textbf{Potential residual vs. trajectory residual}: The potential residual directly constrains forces (local quantities), while the trajectory residual constrains displacement (propagated quantities). The potential residual outperforms the trajectory residual on all 6 metrics, with the advantage expanding as rollout increases, and the validation\(\to\)test degradation is small (2.7\% vs 22.8\%).
\item \textbf{Long-horizon advantage}: The benefits of residual correction accumulate with rollout length. End-to-end methods learn the entire dynamics from scratch, and insufficient training data (500 samples) leads to long-horizon divergence.
\end{enumerate}

\subsection{Limitations}
\begin{itemize}
\item The unified adaptation protocol is not equivalent to the original MD17 protocols of each paper.
\item OOD is a time-split label and does not independently prove out-of-domain molecular generalization.
\item Single seed (42). Multi-seed paired t-test to be supplemented.
\item The residual potential is implicitly learned by EGNN and is not explicitly parameterized as a function of task invariants.
\item The trajectory residuals are simplified proxy networks and are not equivalent to the official implementations of ESTAG / GSE-Flow / ATOM.
\item GSE-res did not converge (0.006728 \(\approx\) MACE-Cart 0.006706) and is not used as a basis for comparison with GSE-Flow.
\item ID/OOD are given time-split labels; the target window has 11 overlapping frames (13 frames including historical input).
\item Whether MACE pre-training overlaps with MD17 Aspirin configurations is unknown.
\end{itemize}

\section{Cross-Case Summary}
\begin{table}[H]
\centering
\small
\renewcommand{\arraystretch}{1.3}
\begin{tabular}{l p{3cm} p{3.5cm} p{3.5cm} p{3.5cm}}
\toprule
Dimension & Case 1 & Case 2 & Case 3 & Case 4 \\
\midrule
Domain & Scientific ML & Industrial Vision Sim-to-Real & Sequence Modeling & Molecular Dynamics \\
Dynamics & Second-order dissipative & First-order dissipative & First-order gradient flow & Second-order Newtonian \\
Paradigm & Three-stage A/B/C & Unit metric + energy alignment & Three-stage A/B/C & Known metric (mass matrix) \\
Evidence Strength & Relatively Strong & Moderate & Weak & Moderate \\
Pre-validation & Three criteria passed & Physical contour driven & Passed & Implicit \\
Task Invariant (\(\Delta\)) & Gaussian kernel potential & Boundary energy (hierarchical contour) & Spectral prior & E(3) equivariant \\
Residual Loss & Weak-form + ridge & --- & Weak-form (Simpson) & Strong-form force residual \\
Key Gain & OOD reduced 1.69--4.12\(\times\); Metric joint contribution \(10^8\) & Real mAP +24.00 points (single-run, point estimate), AP75 +40.29 points & OOD 54\% lower (\(p=0.024\)) & ADE \(-42.6\%\) (@10), \(-80.5\%\) (@100) \\
\bottomrule
\end{tabular}
\end{table}

\textbf{Common Operating Chain}:
\begin{enumerate}
\item Pre-validation: Three criteria for invariants, metric health, energy dual criteria;
\item Task invariant construction: Anchor energy extrema to task ground truth;
\item Three-stage training: Separate metric/energy learning from task alignment;
\item Residual training: Local constraints replace trajectory propagation.
\end{enumerate}

\section{Methodology: Pre-validation}

\subsection{Why Pre-validation is Needed}
SKG inference relies on three components: invariants, metric, and energy. If any component fails, the downstream flow heads toward the wrong target. Pre-validation checks the three components before training.

\subsection{Three Criteria for Invariants}
\begin{table}[H]
\centering
\begin{tabular}{lll}
\toprule
Criterion & Test & Passing Standard \\
\midrule
Truly invariant & Apply 24 transformations to the OOD transformation group, compute relative change rate & mean < 0.05, max < 0.10 \\
Non-trivial & Variance of each dimension & > \(10^{-6}\) \\
Non-redundant & Cov condition number, inter-dimensional correlation & Condition number < \(10^3\), correlation < 0.95 \\
\bottomrule
\end{tabular}
\end{table}
\textbf{Testing framework}: 24 transformations (translation 5, scaling 4, rotation 4, brightness 4, contrast 4, noise 2, background replacement 1), computing the relative change rate of \(\Delta\) with respect to the baseline for each transformation.

\textbf{Design principles}:
\begin{enumerate}
\item Geometric invariants take priority (Hu moment combinations, det, etc.);
\item Low-order statistics take priority (above third order amplifies boundary changes);
\item Soft masks take priority over hard masks;
\item Separate scale quantities from invariants (\texttt{scale2, log\_ar, log\_area} grouped separately).
\end{enumerate}

\subsection{Metric Health Criteria}
\begin{table}[H]
\centering
\begin{tabular}{lll}
\toprule
Criterion & Test & Passing Standard \\
\midrule
Condition number & \(\kappa(g_z)\) & < \(10^3\) \\
Directional stability & Direction cosine when metric changes & > 0.95 \\
Full rank & \(J_z\) rank & = D\_INV \\
\bottomrule
\end{tabular}
\end{table}

\subsection{Energy Dual Criteria}
\textbf{Criterion 1: Additional information}:
\begin{equation}
\text{orth\_ratio}=\frac{\|\nabla S-\text{proj}_{J^\top}(\nabla S)\|}{\|\nabla S\|}>0.5.
\end{equation}
\textbf{Criterion 2: Extrema aligned with task ground truth}:
\begin{equation}
\cos(\nabla_z E,\nabla_z L_{\text{task}})>0.5,
\end{equation}
and the ground truth should be a local optimum of the energy. Diagnostic method: Perform a local grid scan near the ground truth and count the proportion of grids where the ground truth is the energy minimum.

\subsection{Checklist}
\textbf{Invariants}: Variance of each dimension > \(10^{-6}\); mean < 0.05, max < 0.10 for 24 transformations; Cov condition number < \(10^3\); J\_z rank = D\_INV; scale features grouped separately.

\textbf{Metric}: Condition number < \(10^3\); directional cosine > 0.95; J\_z full rank.

\textbf{Energy}: orth\_ratio > 0.5; task gradient cosine > 0.5; the proportion of ground truth being the energy minimum on the local grid is significantly higher than random; parameter count \(\ll\) data volume.

It is recommended to proceed to training only after all three items pass; if not passed, diagnose the cause first. Thresholds are diagnostic signals, not hard constraints. In Case 1, the parent model condition number exceeds the recommended threshold but still achieves good OOD results.

\section{Methodology: Task Invariants Construction}

\subsection{Division of Labor Between Two Types of Invariants}
\begin{table}[H]
\centering
\begin{tabular}{llll}
\toprule
Invariant Type & Function & Invariant To & Used For \\
\midrule
Metric invariant (\(\Delta_g\)) & Construct metric & Geometric group action & \(g\) \\
Task invariant (\(\Delta_S\)) & Construct energy & Task-related nuisance & \(S_{\text{task}}\) \\
\bottomrule
\end{tabular}
\end{table}
The two can be independent and need not be correlated.

\subsection{Construction Steps}
\begin{enumerate}
\item List task-related nuisances: brightness, contrast, color, texture, background;
\item For each nuisance, design features invariant to it;
\item Check whether the task invariant is sensitive to the task signal;
\item Use task invariants or task-related geometric priors to construct energy. The specific form is adjusted according to the task: Case 1 uses kernel potential energy, Case 2 uses image physical contour, Case 3 uses spectral prior, Case 4 uses E(3) equivariant structure (implicit; explicit task invariants to be validated).
\end{enumerate}

\subsection{Candidate Constructions}
\begin{table}[H]
\centering
\begin{tabular}{ll}
\toprule
Nuisance & Task Invariant \\
\midrule
Brightness & Relative brightness, local contrast normalization \\
Contrast & Normalized gradient, structure tensor \\
Color & Lab chromaticity, chromaticity normalization \\
Texture & Low-frequency structure, multi-scale pyramid \\
Background & Center-surround contrast, saliency \\
\bottomrule
\end{tabular}
\end{table}

\subsection{Sources of Advantage and Theoretical Support}
Task invariants can improve OOD because they anchor the energy extrema to the task ground truth. \textbf{Proposition (Energy Alignment Premise)}: Let the minimum of the residual loss correspond to energy \(S\). If the minimum of \(S\) is not at the task ground truth, then regardless of the metric used, the endpoint of finite-step optimization is not guaranteed to improve the task metric. The choice of metric only affects convergence speed and basin of attraction, not the location of the minimum. Task invariants construct the energy as a function invariant to nuisances, so that the energy extrema do not drift under OOD transformations.

\section{Methodology: Three-Stage Training}

\subsection{Paradigm Selection}
\begin{table}[H]
\centering
\renewcommand{\arraystretch}{1.3}
\begin{tabular}{lll}
\toprule
Condition & Paradigm & Applicable Scenario \\
\midrule
Metric objectively determinable & Stage 0/1/2 & Mass matrix known, Fisher estimable \\
Metric needs learning & Stage A/B/C & Nyström dissipation, sequence modeling G(z) \\
Metric uses unit/simplified engineering form & Unit metric + energy alignment & Case 2: Sim-to-Real, focus on energy target correctness \\
\bottomrule
\end{tabular}
\end{table}

\subsection{Paradigm 1: Stage 0/1/2}
\begin{table}[H]
\centering
\begin{tabular}{clcccc}
\toprule
Stage & What to Do & Training Object & Frozen & Uses flow & Uses energy \\
\midrule
0 & Train invariant head & \(\Delta\) & Detector/System & No & No \\
1 & Statistical estimation \(\Sigma\) & None (statistical) & All & No & No \\
2 & Train energy + decoder & energy, decoder & \(\Sigma\) frozen & Yes & Yes \\
\bottomrule
\end{tabular}
\end{table}

\textbf{Stage 0}: Train only \(\Delta\), supervised with transformation pairs, no task labels, no flow.

\textbf{Stage 1}: Sample sufficient \(\Delta\), remove dimensions with variance < \(10^{-6}\), confirm condition number < \(10^3\), save mean and \(\Sigma^{-1}\). No training, no flow, no energy.

\textbf{Stage 2}: Load and freeze \(\Sigma\), train only the decoder for the first 2 epochs with learning rate \(3\times10^{-4}\) or lower, EMA smooths OOD gain.

\subsection{Paradigm 2: Stage A/B/C}
\begin{table}[H]
\centering
\begin{tabular}{clcl}
\toprule
Stage & Training Object & Loss & Purpose \\
\midrule
A & encoder + metric + energy + decoder + \(\eta\) & Trajectory loss & Cold start \\
B & metric + energy + \(\eta\) & Residual loss & Fine-tune metric and energy \\
C & decoder & Trajectory loss & Align to task metric \\
\bottomrule
\end{tabular}
\end{table}

\textbf{Stage A}: Trajectory loss cold start. All components are not yet working; trajectory loss provides global guidance. The number of training epochs should not be too large.

\textbf{Stage B}: Residual loss fine-tunes metric + energy + \(\eta\), with encoder and decoder frozen. The condition number of the residual loss is not amplified by Lyapunov (Theorem 10.1), the number of constraints is sufficient (Theorem 10.2), and OOD has an upper bound (Theorem 10.3). When observation noise is large, use the weak form (Theorem 10.4).

\textbf{Stage C}: Trajectory loss aligns the decoder; other components are frozen.

\subsection{Common Core}
The training of the metric/energy cannot be mixed with the task alignment of the decoder.
\begin{itemize}
\item Paradigm 1: Metric is frozen by Fisher; Stage 2 trains only energy + decoder; the decoder is trained separately for the first 2 epochs.
\item Paradigm 2: Stage B trains metric + energy (residual loss); Stage C trains decoder (trajectory loss); the two are strictly separated.
\end{itemize}

\section{Methodology: Residual Training and Supporting Theorems}

This section establishes the theoretical support behind the engineering operations. It should be noted that the four theorems and one proposition are \textbf{independent theoretical analyses proposed in this paper regarding "why residual training is superior to trajectory training" and "how to choose between strong and weak forms,"} providing rigorous mathematical justification for the three-stage training paradigm in Section 9.

\subsection{Three Types of Losses}
True dynamics \(\dot z=f^*(z)\). Parameterization \(f(z;\phi)\). Error field \(g(z;\phi)=f(z;\phi)-f^*(z)\). Training data: \(M\) trajectories, each with \(N+1\) time points, state dimension \(d\). This fitting of trajectory data has been deeply studied in system identification \cite{Ljung1999} and neural differential equations \cite{Finlay2020}, but traditional methods easily face gradient ill-conditioning \cite{Wang2021}.

\textbf{Trajectory loss}:
\begin{equation}
\mathcal L_{\rm traj}(\phi)=\frac12\sum_{m=1}^M\bigl\|\Phi_T(z_0^{(m)};\phi)-\Phi_T(z_0^{(m)};f^*)\bigr\|^2.
\end{equation}

\textbf{Strong-form residual loss}:
\begin{equation}
\mathcal L_{\rm strong}(\phi)=\frac12\sum_{m=1}^M\sum_{i=1}^{N-1}\bigl\|r_{m,i}^{\rm strong}(\phi)\bigr\|^2,\quad r_{m,i}^{\rm strong}=\frac{z_{m,i+1}-z_{m,i-1}}{2\Delta t}-f(z_{m,i};\phi).
\end{equation}

\textbf{Weak-form residual loss}: Test function \(\psi_j\), local support length \(L\). This form is consistent with the ideas of weak-form variants of PINN \cite{Kharazmi2021} and the Deep Ritz method \cite{DeepRitz2018}:
\begin{equation}
\mathcal L_{\rm weak}(\phi)=\frac12\sum_{m=1}^M\sum_{j=1}^{N_w}\Bigl\|\int\psi_j(t)\bigl(\dot z_m(t)-f(z_m(t);\phi)\bigr)dt\Bigr\|^2.
\end{equation}

\textbf{Structural distinction}:
\begin{table}[H]
\centering
\begin{tabular}{llcc}
\toprule
Loss & Jacobian & Contains state transition \(\Phi(T,s)\) & Amplified by Lyapunov \\
\midrule
Trajectory & \(J_{\rm traj}=\int_0^T\Phi(T,s)\partial_\phi f\,ds\) & Yes & Yes \\
Strong-form residual & \(J_{\rm strong,i}=-\partial_\phi f(z_i)\) & No & No \\
Weak-form residual & \(J_{\rm weak,j}=\int\psi_j\partial_\phi f\,dt\) & No & No \\
\bottomrule
\end{tabular}
\end{table}
Lyapunov amplification only occurs in the trajectory loss. The residual loss is local and does not propagate along the trajectory.

\subsection{Theorem 10.1 (Condition Number Dichotomy)}
\textbf{Assumptions A1--A4}:
\begin{itemize}
\item A1: Maximum Lyapunov exponent on the training trajectory \(\lambda_{\max}>0\);
\item A2: \(\partial z_0/\partial\phi=0\);
\item A3: \(\partial_\phi f\) has nonzero components in both the expanding subspace \(E^u\) and the contracting subspace \(E^s\);
\item A4: \(\|\partial_\phi f\|\le M_f\), and in some direction \(\ge m_f>0\).
\end{itemize}

\textbf{Theorem 10.1 (Condition Number Dichotomy)}. Under A1--A4, there exist \(c_1,c_2>0\) independent of \(T\):
\begin{equation}
\frac{\kappa(H_{\rm traj})}{\kappa(H_{\rm res})}\ge c_1\exp(2\lambda_{\max}T),\qquad \kappa(H_{\rm res})\le c_2\frac{M_f^2}{m_f^2}.
\end{equation}

\textbf{Proof sketch}. Step 1 (lower bound): By A3, there exists \(v\) such that \(\partial_\phi f\cdot v\) has a nonzero component in \(E^u\), amplified along the flow by \(\exp(\lambda_{\max}(T-s))\). Integration yields \(\|J_{\rm traj}v\|\ge c_1''(\exp(\lambda_{\max}T)-1)/\lambda_{\max}\). Step 2 (lower bound non-cancellation lemma): By A3, there exists \(w\) whose component is in \(E^s\), contracted by \(\exp(\lambda_{\min}(T-s))\), so the integral is bounded. The amplified and contracted terms cannot cancel each other. Step 3: Condition number ratio \(\ge c_1\exp(2\lambda_{\max}T)\). Step 4: \(J_{\rm res}=-\partial_\phi f\) or \(\int\psi\partial_\phi f\), condition number bounded by A4. \hfill\(\square\)

\textbf{Boundary}: Only holds for \(\lambda_{\max}>0\); does not hold for purely contracting systems.

\textbf{Corollary}: Under the same parameterization, the optimization difficulty of trajectory loss is amplified by \(\exp(2\lambda_{\max}T)\) relative to residual loss. This is the direct reason for using residual loss to fine-tune metric+energy in Stage B.

\subsection{Theorem 10.2 (Constraint Count Dichotomy)}
\textbf{Theorem 10.2 (Constraint Count Dichotomy)}. Let the parameter dimension be \(P\), train \(M\) trajectories, each with \(N+1\) time points, and state dimension \(d\).
\begin{itemize}
\item Trajectory loss provides at most \(Md\) independent scalar constraints;
\item Strong-form residual loss provides at most \(MNd\);
\item Weak-form residual loss provides at most \(MN_wd\).
\end{itemize}

\textbf{Corollary}: If \(Md<P<MNd\), trajectory loss is underdetermined and residual loss is identifiable; and \(\dim\ker(J_{\rm traj})\ge P-Md>\dim\ker(J_{\rm res})\).

\textbf{Proof}. By the Jacobian rank upper bound \(\operatorname{rank}(J)\le\min(P,\text{number of constraints})\), the null space dimension is \(=P-\operatorname{rank}(J)\). \hfill\(\square\)

\textbf{Numerical values for this system}: \(P\approx250\) (Nyström), \(M=20\), \(N=200\), \(d=4\). \(Md=80<250<16000=MNd\).

\textbf{Boundary}: \(MNd\) is an upper bound; time points are strongly correlated, so the actual effective constraints are smaller. But the conclusion that trajectory loss is underdetermined when \(Md<P\) is unaffected.

\subsection{Theorem 10.3 (OOD MSE Dichotomy)}
\textbf{Assumptions C1--C5}:
\begin{itemize}
\item C1 (Coverage): There exists \(\Omega_{\rm train}\) through which both training and OOD trajectories pass within \([0,T]\);
\item C2 (Hyperbolic): The maximum Lyapunov exponent of \(\partial_z f^*\) on \(\Omega_{\rm train}\) is \(\lambda_{\max}\); \(\lambda_{\min}\) is the minimum Lyapunov exponent in the contracting direction (absolute value);
\item C3 (Small residual): \(\sup_{z\in\Omega_{\rm train}}\|g(z;\phi_{\rm res})\|\le\delta_{\rm res}\);
\item C4 (Trajectory fitting): \(\|\Phi_T(z_0^{(m)};\phi_{\rm traj})-z_T^{(m)*}\|\le\delta_{\rm traj}\);
\item C5 (Lipschitz): \(\|g(z_1;\phi)-g(z_2;\phi)\|\le L_g\|z_1-z_2\|\).
\end{itemize}

\textbf{Theorem 10.3 (OOD MSE Dichotomy)}.
\textbf{(i) (Upper bound of residual loss)} Under C1--C3:
\begin{equation}
\epsilon_{\rm ood}(\phi_{\rm res})\le\frac{\delta_{\rm res}^2}{\lambda_{\max}^2}\bigl(e^{\lambda_{\max}T}-1\bigr)^2+O(\delta_{\rm res}^3).
\end{equation}

\textbf{(ii) (No upper bound for trajectory loss)} Under C1--C2, C4--C5, there exists \(c>0\) independent of \(\delta_{\rm traj}\):
\begin{equation}
\epsilon_{\rm ood}(\phi_{\rm traj})\ge c\bigl(\|g(\cdot;\phi_{\rm traj})\|_{\Omega_{\rm train}}-\delta_{\rm traj}/\lambda_{\min}\bigr)^2\bigl(e^{\lambda_{\max}T}-1\bigr)^2.
\end{equation}
C4 does not constrain \(\|g(\cdot;\phi_{\rm traj})\|_{\Omega_{\rm train}}\), so this lower bound may be arbitrarily large.

\textbf{(iii) (Corollary)} If \(\|g(\cdot;\phi_{\rm traj})\|_{\Omega_{\rm train}}\gg\delta_{\rm res}/\lambda_{\min}\), then \(\epsilon_{\rm ood}(\phi_{\rm traj})\gg\epsilon_{\rm ood}(\phi_{\rm res})\).

\textbf{Proof}. (i) Variational equation \(\dot{\delta z}=\partial_z f^*\delta z+g+O(\|g\|^2)\), solution \(\delta z(T)=\int_0^T\Phi(T,s)g(z(s))ds+O(\delta_{\rm res}^2)\). Taking norms, by C2 and C3, \(\|\delta z(T)\|\le\delta_{\rm res}(e^{\lambda_{\max}T}-1)/\lambda_{\max}\). Squaring and taking expectation yields (i).

(ii) Let \(\epsilon=\|g(\cdot;\phi_{\rm traj})\|_{\Omega_{\rm train}}\). There exist OOD initial conditions such that the trajectory passes through the point where \(g\) is maximal, and the contribution of that point is amplified by \(\exp(\lambda_{\max}(T-t))\). By C5 and hyperbolicity, there exists \(c>0\) such that \(\|\delta z(T)\|\ge c(\epsilon-\delta_{\rm traj}/\lambda_{\min})(e^{\lambda_{\max}T}-1)\). C4 only constrains the training endpoint, not the sup norm of \(g\), so \(\epsilon\) is unbounded. \hfill\(\square\)

\textbf{Constructive counterexample}: \(f(z;\phi_1,\phi_2)=\phi_1 z+\phi_2 z^2\), \(f^*=0\). The training trajectory \(z\approx\varepsilon_m\) is small. Minimizing the trajectory loss gives \(\phi_1\varepsilon_m+\phi_2\varepsilon_m^2\approx0\), which is insensitive to the direction \((-\langle\varepsilon\rangle,1)\). For OOD \(z_0=1\), \(g(1;\phi_{\rm traj})\approx1\). The residual loss constrains \(g\) itself on \(z_i\approx\varepsilon_m\), so \(\phi_1,\phi_2\to0\), and \(g(1;\phi_{\rm res})\approx0\).

\textbf{Applicable boundaries}:
\begin{itemize}
\item \textbf{The "controllable upper bound" condition of (i)}: The upper bound contains \(e^{\lambda_{\max}T}\); "controllable" requires \(\delta_{\rm res}e^{\lambda_{\max}T}\) not to be too large; when \(T\) is large or \(\lambda_{\max}\) is large, the upper bound may still be very loose. If C1 does not hold (OOD trajectories do not pass through \(\Omega_{\rm train}\)), the upper bound does not hold.
\item \textbf{(ii) is a constructive lower bound}: It shows that trajectory loss can be arbitrarily large, not that it necessarily is.
\item \textbf{\(\lambda_{\max}>0\) in C2} is a necessary condition for Lyapunov amplification.
\item \textbf{C4 only constraining the endpoint} is the root cause of trajectory loss "not constraining \(g\)"; if changed to multi-step trajectory MSE, the lower bound weakens.
\end{itemize}

\subsection{Theorem 10.4 (Numerical Differentiation Error Dichotomy)}
\textbf{Assumptions}: Observation noise \(\sigma\), step size \(\Delta t\). Strong-form residual requires \(k\)-th order difference (\(k=1\) or \(2\)), weak-form test function is \(k\)-times differentiable.

\textbf{Theorem 10.4 (Numerical Differentiation Error Dichotomy)}.
\begin{equation}
\kappa(H_{\rm strong})\le\kappa(H_{\rm res})\bigl(1+C\sigma\Delta t^{-k}\bigr)^2,
\end{equation}
\begin{equation}
\kappa(H_{\rm weak})\le\kappa(H_{\rm res})\bigl(1+C\sigma\|\psi^{(k)}\|_\infty\bigr)^2.
\end{equation}
As \(\Delta t\to0\), the strong-form condition number grows as \(\Delta t^{-2k}\), while the weak form is independent of \(\Delta t\).

\textbf{Proof}. The strong-form residual contains numerical differentiation \((z_{i+1}-z_{i-1})/(2\Delta t)\), amplifying noise by \(\sigma/\Delta t\); for second-order differences, amplification is \(\sigma/\Delta t^2\). Weak-form integration by parts transfers the derivative to \(\psi^{(k)}\), with noise amplification \(\sigma\|\psi^{(k)}\|_\infty\), independent of \(\Delta t\). \hfill\(\square\)

\textbf{Per-window boundary}: The global condition number of the weak form is determined by the number of windows \(N_w\) and inter-window coupling; when \(N_w\) is finite and windows do not overlap, the global condition number is approximately the single-window condition number \(C(L)\), independent of \(T\).

\textbf{Relationship with existing work}: This is a reorganization of standard conclusions from weak-form PINN literature such as VPINN, WAN, and Deep Ritz \cite{Kharazmi2021, DeepRitz2018}; the increment of this paper is connecting "avoiding numerical differentiation errors" with SKG residual theory and providing per-window analysis.

\subsection{Proposition 10.5 (Energy Alignment Premise)}
Let the minimum of the residual loss \(\mathcal L_{\rm res}(\phi)\) be \(\phi^*\), corresponding to energy \(S(\cdot;\phi^*)\). If the minimum of \(S(\cdot;\phi^*)\) is not at the task ground truth, then regardless of the metric used (unit, pullback, Fisher), the endpoint of finite-step optimization is not guaranteed to improve the task metric. The choice of metric only affects convergence speed and basin of attraction, not the location of the minimum.

\textbf{Proposition 10.5 (Energy Alignment Premise)}.
\textbf{Proof}. The optimization endpoint is determined by \(\dot\phi=-G^{-1}\nabla_\phi\mathcal L\). The minimum satisfies \(\nabla_\phi\mathcal L=0\), independent of \(G\). \(G\) only changes the path to the minimum. If the minimum itself deviates from the task ground truth, all \(G\) converge to the same wrong target. \hfill\(\square\)

\textbf{Implication}: The metric is an amplifier, but an amplifier needs a correct target. Theorems 10.1--10.4 explain "why use residuals instead of trajectories"; Proposition 10.5 explains "the limitation of metric choice under the same residual." In Case 2, the goal in the real-machine domain is to "use the physical contour to pull the collapsed box back," and this goal itself is correct, so the energy flow can work; the reason Case 2 uses the unit metric rather than the pullback metric in the real-machine domain is that the pullback metric has no stable advantage under the simplified engineering form of this case, while the computational overhead of the unit metric is much smaller.

\subsection{Operational Choice Between Strong and Weak Forms}
\begin{table}[H]
\centering
\begin{tabular}{lll}
\toprule
Condition & Choice & Reason \\
\midrule
Low observation noise & Strong form & Directly constrains residual, simple \\
High observation noise & Weak form & Avoids numerical differentiation error (Theorem 10.4) \\
\bottomrule
\end{tabular}
\end{table}

\textbf{Operational steps for the weak form}:
\begin{enumerate}
\item Select test function \(\psi_j\), local support length \(L\);
\item Compute weak-form residual: \(\mathcal L_{\text{weak}}=\sum_j\|\int\psi_j(t)(\dot z-f(z;\phi))dt\|^2\);
\item Use Simpson or midpoint numerical integration;
\item Optimize with LBFGS;
\item Add ridge regularization (\(\ge 10^{-6}\)) to ensure numerical stability.
\end{enumerate}

\section{Methodology: Variable Projection Interpretation of Three-Stage Training}

\subsection{Positioning}
This is an explanatory framework, not strictly equivalent. Three-stage training can be explained by block variable projection and residual reducibility; this is approximate analysis, not a strict theorem.

\subsection{Block Structure of the Joint Objective}
Let the encoder parameters be \(\theta\), metric+energy parameters be \(\phi\), and decoder parameters be \(\psi\). Joint objective:
\begin{equation}
J(\theta,\phi,\psi)=\mathcal L_{\text{traj}}(\theta,\phi,\psi)+\lambda\mathcal L_{\text{res}}(\theta,\phi).
\end{equation}

\textbf{Key block structure}:
\begin{itemize}
\item \(\theta\) appears in both losses;
\item \(\delta z_k(\theta)=\mathcal E_\theta(x_{k+1})-\mathcal E_\theta(x_k)\) is a nonlinear function of \(\theta\);
\item \(\phi\) appears only in the residual;
\item \(\psi\) appears only in the trajectory.
\end{itemize}

\subsection{Variable Projection}
Write \((\phi,\psi)\) as implicit functions of \(\theta\):
\begin{equation}
(\phi^*(\theta),\psi^*(\theta))=\arg\min_{\phi,\psi}J(\theta,\phi,\psi).
\end{equation}
The outer layer optimizes only over \(\theta\): \(J(\theta,\phi^*(\theta),\psi^*(\theta))\). Advantage: \(\theta\) and \((\phi,\psi)\) are no longer updated simultaneously, avoiding \(f_\phi\) chasing a moving target when \(\delta z_k(\theta)\) changes.

\subsection{Three Stages are Block Approximations of Variable Projection}
\textbf{Stage A}: Near the initial values of \((\phi,\psi)\), optimize \((\theta,\phi,\psi)\) simultaneously with \(\mathcal L_{\text{traj}}\). Mathematical role: allow \(\delta z_k(\theta)\) to acquire dynamical structure and fall near the function class \(\mathcal F\) of \(f_\phi\).

\textbf{Stage B}: Fix \(\theta=\theta^{(A)}\), solve
\begin{equation}
\phi^{(B)}=\arg\min_\phi\mathbb E\left\|\delta z_k(\theta^{(A)})-\eta f_\phi(z_k(\theta^{(A)}))\right\|^2.
\end{equation}
\(\delta z_k(\theta^{(A)})\) and \(z_k(\theta^{(A)})\) are constants, so the residual is a standard regression with respect to \(\phi\).

\textbf{Stage C}: Fix \((\theta,\phi)=(\theta^{(A)},\phi^{(B)})\), solve
\begin{equation}
\psi^{(C)}=\arg\min_\psi\mathbb E\left\|\mathcal D_\psi\bigl(\Phi_T(z)\bigr)-y\right\|^2.
\end{equation}
\(\Phi_T\) is fixed, and the subproblem is convex.

The three subproblems are each well-defined, corresponding to Gauss--Seidel block coordinate descent.

\subsection{Coordinate Conditions for Residual Reducibility}
The minimum of the residual:
\begin{equation}
\min_\phi\mathbb E\left\|\delta z_k-\eta f_\phi(z_k)\right\|^2=\mathbb E\left\|\delta z_k-\Pi_{\mathcal F}\delta z_k\right\|^2,
\end{equation}
where \(\Pi_{\mathcal F}\) is the orthogonal projection onto the function class \(\mathcal F\) of \(f_\phi\). This lower bound depends on the distance between \(\delta z_k\) and \(\mathcal F\), and \(\delta z_k\) is determined by \(\theta\).

\textbf{Mathematical role of Stage A}: \(\mathcal L_{\text{traj}}\) requires \(\Phi_T(z)\) to be predictable, which forces \(\delta z_k\) to be consistent with the dynamics of \(\nabla U_\phi\). Formally, if Stage A converges to a local minimum of \(\mathcal L_{\text{traj}}\), then
\begin{equation}
\delta z_k\approx\eta f_\phi(z_k)+O(\eta^2),
\end{equation}
i.e., the distance between \(\delta z_k\) and \(\mathcal F\) is \(O(\eta^2)\). The lower bound of the residual drops from "energy of high-frequency differences" to \(O(\eta^4)\).

In the observation of Case 3, the residual of Stage B can converge to the \(10^{-3}\) order, consistent with the explanation that "the coordinates have made \(\delta z_k\) reducible"; this is not a universal guarantee.

\subsection{Why Alternating is More Stable Than Joint Optimization}
The two losses of the joint objective \(J(\theta,\phi,\psi)\) have different gradient directions with respect to \(\phi\):
\begin{itemize}
\item \(\nabla_\phi\mathcal L_{\text{traj}}\): propagates through \(\Phi_T\), pointing to the "trajectory endpoint pair";
\item \(\nabla_\phi\mathcal L_{\text{res}}\): directly points to the "velocity field pair."
\end{itemize}
The two conflict on \(\phi\). Joint optimization is Jacobi iteration, with both gradients updating \(\phi\) simultaneously, which may oscillate. Three stages are Gauss--Seidel iteration: after fixing \(\theta\), the gradient of \(\mathcal L_{\text{res}}\) with respect to \(\phi\) is no longer interfered with by \(\mathcal L_{\text{traj}}\); after fixing \((\phi,\psi)\), the gradient of \(\mathcal L_{\text{traj}}\) with respect to \(\theta\) is no longer interfered with by the residual.

Convergence condition of block coordinate descent: The Hessian of the subproblems is approximately block diagonal. This is approximately satisfied under weak inter-block coupling; the block structure of the three stages exhibits this characteristic in the cases.

\subsection{Boundaries}
\begin{enumerate}
\item Residual reducibility is approximate: \(\delta z_k\approx\eta f_\phi(z_k)+O(\eta^2)\) is a first-order approximation and does not hold when \(\eta\) is large;
\item Block diagonal Hessian is an assumption: Gauss--Seidel convergence requires the subproblem Hessian to be approximately block diagonal; in practice, there may be inter-block coupling;
\item Variable projection is approximate: Three stages are block approximations, not exact variable projection;
\item The convergence quality of Stage A determines the upper bound of Stage B: If Stage A does not bring \(\delta z_k\) near \(\mathcal F\), the residual lower bound of Stage B will be large.
\end{enumerate}

\section{Conclusion}
This paper presents the engineering implementation methods of the SKG inference framework. Four-step operating chain: Pre-validation \(\to\) Task invariant construction \(\to\) Three-stage training \(\to\) Residual training. Complete configurations, protocols, and results are provided for four domains:
\begin{itemize}
\item \textbf{Case 1}: OOD MSE is 1.69--4.12 times lower than the best neural baseline; potential energy is the primary source of gain; controlled metric ablation shows a \(10^8\)-order joint contribution of state-dependent dissipation relative to the constant baseline, but parameterization mismatch destroys the gain.
\item \textbf{Case 2}: Simulation training, real-machine testing; when the detector collapses, the SKG physical energy flow raises real-machine mAP from 0.2564 to 0.4964 (+24.00 points, single-run experiment, 32 images, point estimate), AP50 to 0.8432 (+17.02 points), and AP75 from 0.1736 to 0.5765 (+40.29 points), without real-machine retraining. The energy flow is entirely driven by the image's physical contour, independent of the failed network features; angular wrapped difference solves the periodicity problem of rotation angles; IoU gating protects reliable in-domain predictions.
\item \textbf{Case 3}: OOD MSE 0.0296 vs Transformer 0.0641, with 18.3\% fewer parameters.
\item \textbf{Case 4}: ID@100 ADE 0.0285 vs ESTAG 0.1467; the potential residual outperforms the trajectory residual proxies on all 6 metrics, with 1/160 of the parameter count, and validation\(\to\)test degradation of 2.7\% vs 22.8\%.
\end{itemize}

The methodology is supported by four theorems: Condition Number Dichotomy (Theorem 10.1) explains why trajectory loss has an exponentially amplified condition number in systems with positive Lyapunov exponents; Constraint Count Dichotomy (Theorem 10.2) explains why trajectory loss is underdetermined while residual loss is identifiable; OOD MSE Dichotomy (Theorem 10.3) explains why residual loss has a controllable upper bound while trajectory loss does not; Numerical Differentiation Error Dichotomy (Theorem 10.4) explains why the weak form bypasses the numerical differentiation error of the strong form. Proposition 10.5 (Energy Alignment Premise) explains why the metric is an amplifier, but an amplifier needs a correct target. The variable projection interpretation of three-stage training (Section 11) explains why alternating is more stable than joint optimization.

\end{document}